\documentclass[10pt,twocolumn,letterpaper]{article}

\usepackage{cvpr}      

\usepackage{colortbl}

\usepackage{graphicx}
\usepackage{booktabs}
\usepackage{multirow}
\usepackage{makecell}
\usepackage[table]{xcolor}
\usepackage{array}

\usepackage[percent]{overpic}
\usepackage{microtype}           
\usepackage{nicefrac}            
\usepackage{xspace}              
\usepackage{anyfontsize}
\usepackage{amsmath, amssymb, amsfonts, amsthm, mathtools,mathrsfs}
\usepackage{dsfont}              
\usepackage{fdsymbol}            
\usepackage[ruled,vlined]{algorithm2e}

\usepackage{booktabs}            
\usepackage{nicematrix}         
\usepackage{multirow}
\usepackage{makecell}
\usepackage{bigstrut}
\usepackage{enumitem}            
\setitemize{label=\textbullet, leftmargin=*, nolistsep}
\usepackage{graphicx}
\usepackage{adjustbox}           
\usepackage{float}               
\usepackage{wrapfig}             
\usepackage{subcaption}          
\expandafter\def\csname ver@subfig.sty\endcsname{}  
\usepackage{fontawesome5}        
\usepackage{pifont}              
\usepackage{tikz}

\newcommand{\circnum}[1]{%
  \tikz[baseline=(char.base)]{
    \node[shape=circle,draw,inner sep=0.5pt] (char) {\footnotesize #1};
  }%
}
\definecolor{cvprblue}{rgb}{0.21,0.49,0.74}
\usepackage[pagebackref,breaklinks,colorlinks,allcolors=cvprblue]{hyperref}
\usepackage[all]{hypcap}         
\usepackage{url}                 
\usepackage[normalem]{ulem}      
\usepackage{lipsum}              
\usepackage{rotating}            
\usepackage{stackengine}         
\usepackage{caption}             
\usepackage{listings}            
\usepackage{soul}                
\usepackage{gradient-text}       
\usepackage{pgfplots}            
\pgfplotsset{compat=newest}
\usepackage{pgfplotstable}       
\usepackage{tikz}                
\usepackage{svg}                 

\usepackage{wrapstuff}
\usepackage{makecell}

\usepackage{array}
\usepackage{makecell}
\usepackage{booktabs}

\newcolumntype{x}[1]{>{\centering\arraybackslash}p{#1}}

\definecolor{demphcolor}{RGB}{125,125,125}             

\newcommand{\xmark}{\textcolor{red}{\ding{55}}}     

\newcommand{\logoicon}{%
  \raisebox{-0.20\height}{\includegraphics[height=1.2em]{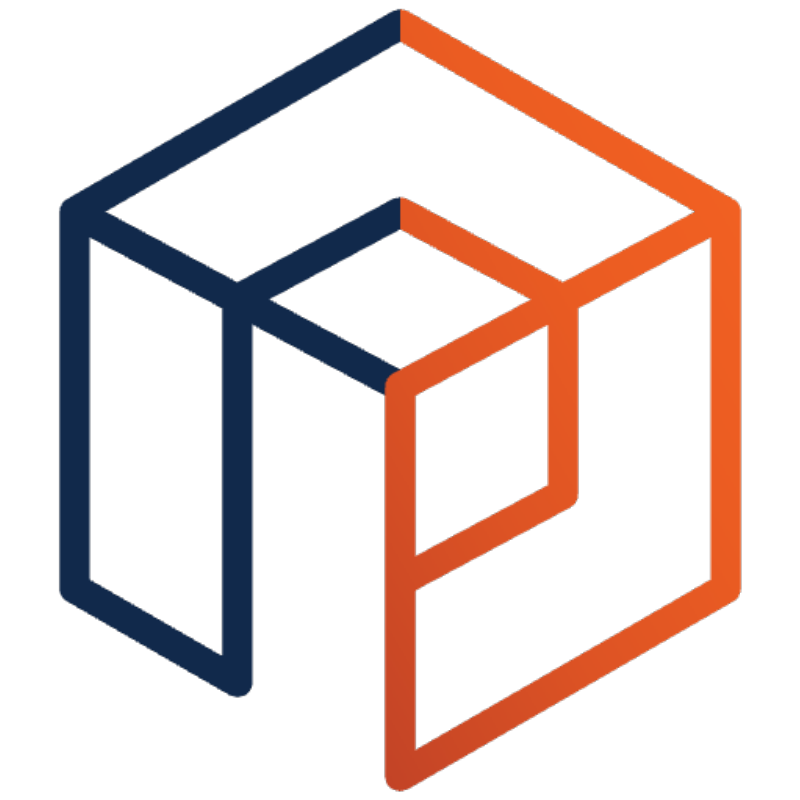}}%
}
\def\paperID{} 
\def\confName{CVPR}
\def\confYear{2026}

\definecolor{IllinoisOrange}{HTML}{FF5F05}
\definecolor{IllinoisBlue}{HTML}{13294B}
\newcommand{\modelname}{\textbf{\textsc{PALM}}\xspace}
\newcommand{\modelnamenc}{{\textsc{PALM}}\xspace}

\title{PALM: Progress-Aware Policy Learning via Affordance Reasoning for Long-Horizon Robotic Manipulation\vspace{-0.3cm}}

\author{
Yuanzhe Liu$^{1,2}$ \quad
Jingyuan Zhu$^{1}$ \quad
Yuchen Mo$^{2}$ \quad
Gen Li$^{3}$ \quad
Xu Cao$^{2}$ \quad
Jin Jin$^{4}$ \quad
Yifan Shen$^{2}$ \\
Zhengyuan Li$^{2}$ \quad
Tianjiao Yu$^{2}$ \quad
Wenzhen Yuan$^{2}$ \quad
Fangqiang Ding$^{5}$ \quad
Ismini Lourentzou$^{2}$ \\[1ex]
$^{1}$University of Pennsylvania \quad
$^{2}$University of Illinois Urbana-Champaign \\
$^{3}$Nanyang Technological University \quad
$^{4}$University of Oxford \quad
$^{5}$Massachusetts Institute of Technology \\[0.5ex]
}

\begin{document}

\twocolumn[{
\maketitle
\begin{center}
\vspace{-0.8cm}
\captionsetup{type=figure}\includegraphics[width=0.97\linewidth]{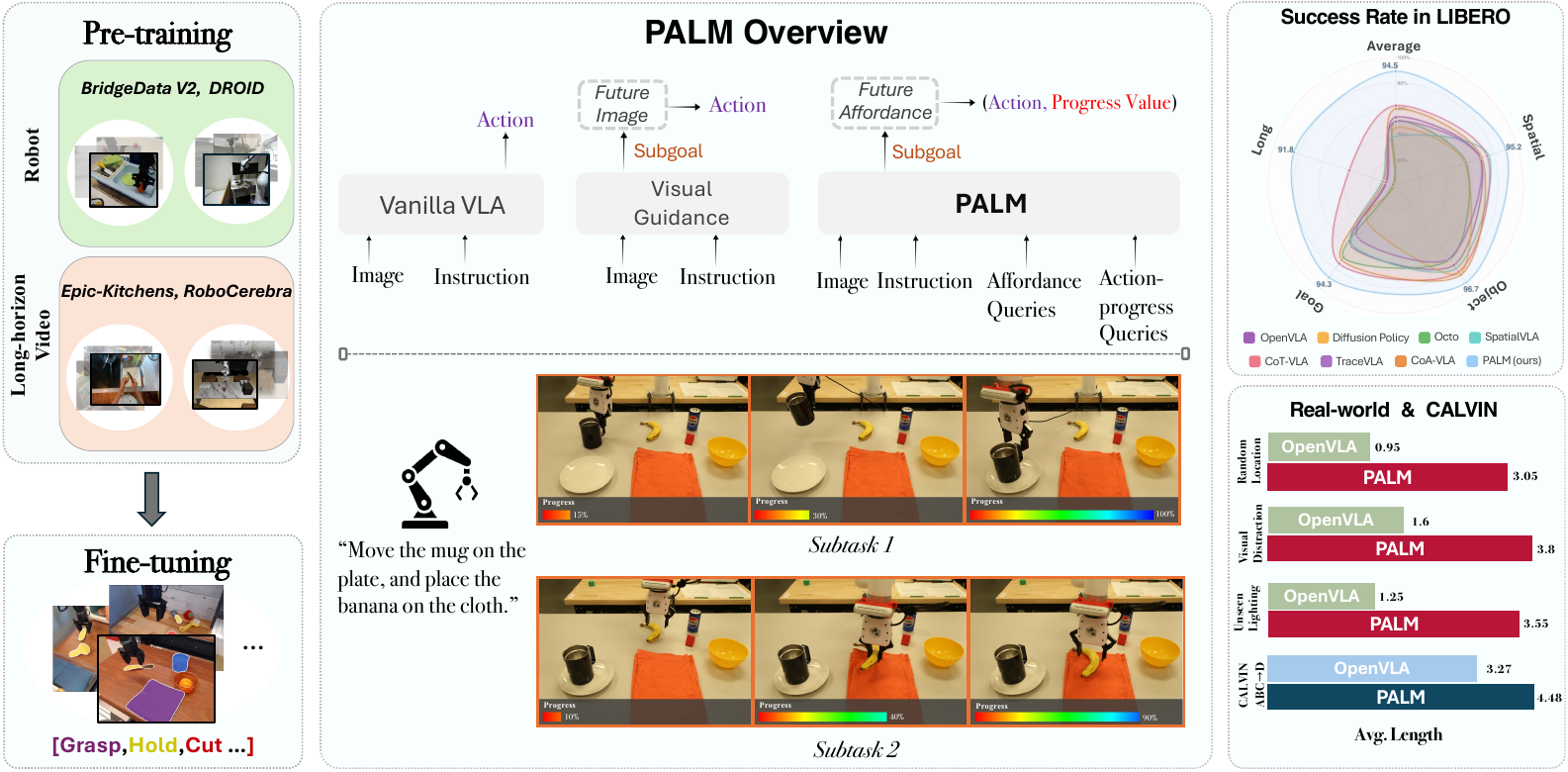}  
    \vspace{-0.3cm}
    \captionof{figure}{In contrast to vanilla VLAs that directly map inputs to actions or to predictive methods that forecast dense future images, \modelname introduces learnable queries to forecast a structured set of future affordances. Conditioned on these affordances, a diffusion-based policy jointly decodes the robot's action and a continuous progress value, enabling temporal state tracking and seamless subtask transitions in multi-step, long-horizon tasks. Our training strategy utilizes large-scale pre-training on both robot datasets (BridgeDataV2~\cite{walke2023bridgedata}, DROID~\cite{khazatsky2024droid}) and long-horizon video data (EPIC-KITCHENS~\cite{damen2020epic}, RoboCerebra~\cite{han2025robocerebra}), followed by fine-tuning on our collected human-annotated affordance dataset. Consequently, \modelname attains state-of-the-art performance on long-horizon simulation benchmarks (CALVIN ABC$\rightarrow$D~\cite{mees2022calvin}, LIBERO-LONG~\cite{liu2023libero}) and demonstrates strong results under real-world long-horizon generalization settings.}
    \label{fig:teaser}
\end{center}
\vspace{0.2cm}
}]

\begin{abstract}
Recent advancements in vision-language–action (VLA) models have shown promise in robotic manipulation, yet they continue to struggle with long-horizon, multi-step tasks. Existing methods lack internal reasoning mechanisms that can identify task-relevant interaction cues or track progress within a subtask, leading to critical execution errors such as repeated actions, missed steps, and premature termination. To address these challenges, we introduce \modelname, a VLA framework that structures policy learning around interaction-centric affordance reasoning and subtask progress cues. \modelnamenc distills complementary affordance representations that capture object relevance, contact geometry, spatial placements, and motion dynamics, and serve as task-relevant anchors for visuomotor control. To further stabilize long-horizon execution, \modelnamenc predicts continuous within-subtask progress, enabling seamless subtask transitions. Across extensive simulation and real-world experiments, \modelnamenc consistently outperforms baselines, achieving a 91.8\% success rate on LIBERO-LONG, a 12.5\% improvement in average length on CALVIN ABC$\rightarrow$D, and a 2$\times$ improvement over real-world baselines across three long-horizon generalization settings.
\noindent \logoicon~\href{https://plan-lab.github.io/palm}{\textcolor{IllinoisBlue}{PLAN Lab}~\textcolor{IllinoisOrange}{https://plan-lab.github.io/palm}}
\end{abstract}

\vspace{-1em}
\section{Introduction}
Robot learning has made significant strides in developing generalizable policies for diverse tasks and environments~\cite{bharadhwaj2024gen2act, cheang2024gr, driess2023palm, du2023video, fu2024mobile, lin2024data, zawalski2024robotic, wen2023any, wu2023unleashing, yang2024pushing, zhang2025learning,hao2025mimo}. Much of this progress is driven by Vision-Language-Action (VLA) models, which leverage pre-trained vision-language backbones to map visual observations and language instructions directly to robot actions~\cite{zhen20243d, team2024octo,kim2024openvla,black2024pi0visionlanguageactionflowmodel,intelligence2025pi05visionlanguageactionmodelopenworld,chi2025diffusion,li2025coa,black2023zero,brohan2022rt,zitkovich2023rt,nasiriany2025rt,gu2023rt,li2023vision,xiao2022masked,zeng2024learning,reuss2024multimodal}. 
However, current VLA methods are fundamentally limited to short-horizon manipulation and struggle with long-horizon, multi-step planning in dynamic scenes. For example, on “clean a cluttered table,” state-of-the-art policies typically succeed initially but fail mid-task, unable to reliably complete the full sequence.\looseness-1

A fundamental limitation is the absence of structured affordance cues~\cite{yuan2024robopoint,kuang2024ram,sundaresan2023kite,huang2024copa,huang2024rekep} and explicit state tracking~\cite{chen2025sarm,kang2025incorporating}. Although existing models may infer the final goal and produce intermediate actions~\cite{zhang2025up,zhang2025dreamvla,hu2024video,tian2024predictive,chen2025villa,zhao2025cot}, they lack internal representations that disambiguate which object should be targeted next, which part or region is relevant for interaction, where items should be placed or moved, or what motion is appropriate for the upcoming step. Consequently, many visually similar states become ambiguous, obscuring the underlying task stage and destabilizing long-horizon control.
In addition, existing VLAs lack mechanisms for continuously estimating progress within a subtask. Without a persistent online notion of advancement, the policy cannot reliably decide whether to continue, switch stages, or terminate. This absence of temporal grounding leads to characteristic long-horizon failure modes: repeated or unnecessary actions, skipped required subtasks, premature termination, and even declaring success in incorrect states.

To address these gaps, we introduce \modelname, a novel end-to-end framework for learning scalable, long-horizon manipulation. As illustrated in \cref{fig:teaser}, \modelnamenc integrates perception, action, and progress within a closed loop and is built around two complementary capabilities. First, \modelnamenc predicts structured future affordances encoded as latent queries, which are refined through a block-wise structured attention mechanism and specialize into four types, capturing object relevance (global), contact-level geometry (local), candidate placement regions (spatial), and plausible motion trajectories for the next interaction step (dynamic). These latents form a compact and task-relevant representation of the evolving scene state. 
Second, \modelnamenc estimates fine-grained subtask progress through a continuous progress signal. Conditioned on both the affordance latents and the current multimodal context, the action policy employs a diffusion transformer to jointly predict action and progress sequences.
By combining structured affordance prediction with progress-aware action generation, \modelnamenc maintains stable, coherent behavior across long, multi-step manipulation sequences.

We conduct large-scale pre-training on both robot~\cite{khazatsky2024droid, walke2023bridgedata} and long-horizon video data~\cite{damen2020epic,han2025robocerebra}, and then fine-tune the model on human-annotated robot trajectories. Through extensive experiments in both simulation and real-world settings, \modelnamenc demonstrates clear and consistent gains in long-horizon manipulation. On two widely used long-horizon simulation benchmarks, it improves the success rate on CALVIN ABC$\rightarrow$D~\cite{mees2022calvin} by 12.5\% over prior state-of-the-art baselines and reaches 91.8\% success on LIBERO-LONG~\cite{liu2023libero}. Beyond simulation, we design three types of challenging real-world generalization tests for long-horizon scenarios that vary in localization, lighting, and visual distractors. Under all conditions, \modelnamenc consistently demonstrates strong performance and robustness with limited fine-tuning data. Our contributions are as follows:
\begin{itemize}
    \item We introduce \modelname, a unified VLA framework that integrates structured affordance reasoning and progress-aware policy generation to enable reliable execution across long-horizon, contact-rich manipulation tasks.
    \item We propose two novel complementary modules: (1) a fine-grained affordance predictor that acts as an intermediate implicit reasoning step, producing structured and task-relevant representations, and (2) a progress-aware inverse-dynamics module that couples action generation with subtask progress estimation, ensuring temporally coherent execution over long action sequences.
    \item We conduct comprehensive evaluations both in simulation and the real world, achieving a 12.5\% improvement on CALVIN ABC$\rightarrow$D over prior state-of-the-art, 91.8\% success on LIBERO-LONG, and consistent gains across three real-world long-horizon generalization tests.
\end{itemize}

\section{Related Work}
\paragraph{Vision-Language-Action Models.} 
Several works repurpose pre-trained Vision-Language Models (VLMs)~\cite{li2025oric,shen2025fine,li2026toward,ahn2022can,huang2022inner,mu2025should,zhu2026can,zhu2026fusionagent, dou2025plan} into VLA policies that map visual observations and language instructions to low-level robot actions~\cite{shridhar2022cliport, deng2025graspvla, zhao2025vlas, karamcheti2023language, driess2023palm,xu2025stare,lin2025evo0,lin2025evo1,ji2025robobrain,bjorck2025gr00t} by fine-tuning on large-scale robotics datasets \cite{o2024open, ebert2021bridge, khazatsky2024droid}. A prominent paradigm, pioneered by the RT series \cite{brohan2022rt, zitkovich2023rt, belkhale2024rt}, formulates action generation as autoregressive prediction over tokenized sequences~\cite{blattmann2023stable, zhu2024vision, li2024llara, kim2024openvla}.
In parallel, diffusion-based action generators \cite{chi2025diffusion, hou2025dita, liu2024rdt, ke20243d, ze20243d}
treat control as a denoising process in continuous trajectory spaces, producing smoother temporal dynamics. Despite their success, both paradigms rely on direct action prediction, which lacks explicit reasoning and fine-grained representations of spatial or physical dynamics. Subsequent works address these limitations by incorporating future prediction through goal-image generation \cite{du2023learning, zawalski2024robotic, zhen20243d, gu2023rt, wen2023any, hu2024video} or integrated forecasting \cite{tian2024predictive, zhang2025up, zhao2025cot, yang2025gripper, zhu2025unified}, or by enhancing spatio-temporal grounding via keypoint prediction \cite{yuan2024robopoint, huang2024rekep} and historical visual traces \cite{niu2024llarva}. In contrast, our work introduces a closed perception–action–progress loop that improves long-horizon manipulation by integrating affordance and subtask progress reasoning into the VLA.

\begin{figure*}[t!]
    \centering
    \includegraphics[width=0.99\linewidth]{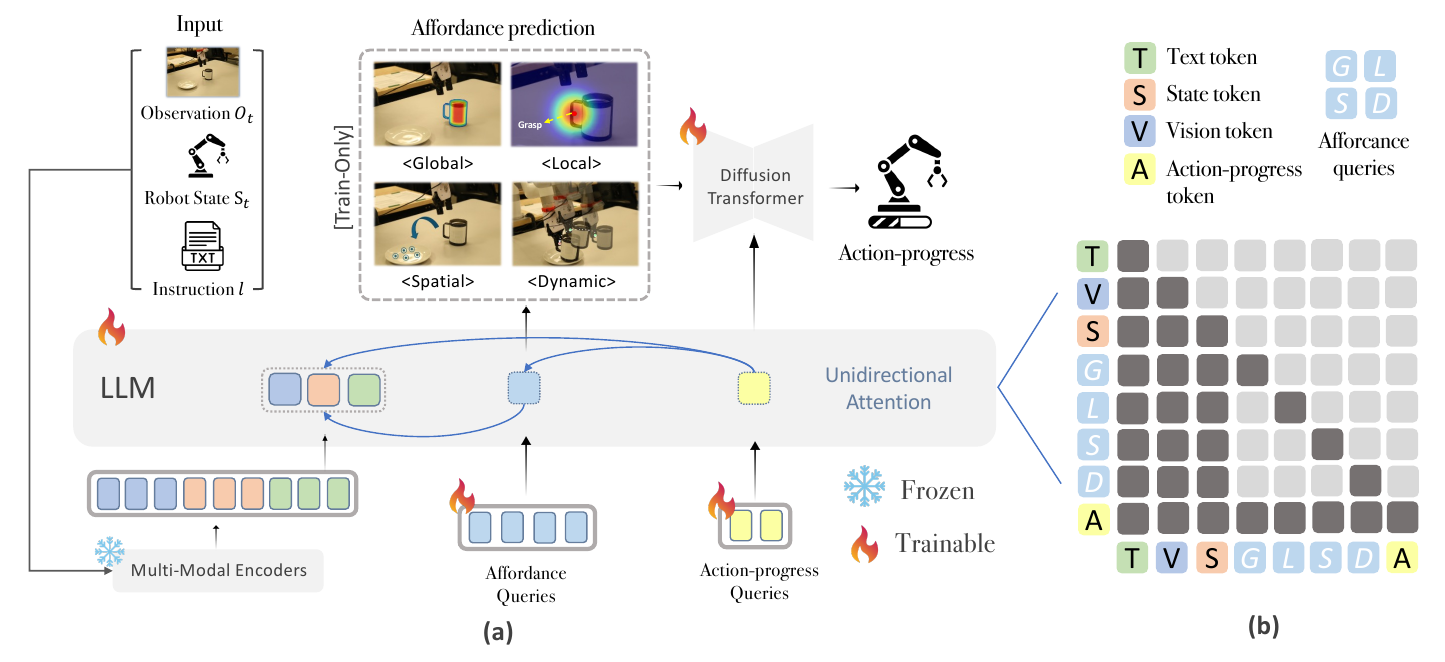}
    \vspace{-0.5cm}
    \caption{
    \textbf{\modelname Overview.}
    \textbf{ (a) Model Architecture:}
    Given a language instruction $l$, observation $o_t$, and robot state $s_t$, \modelnamenc\ encodes each modality using frozen encoders to obtain text, visual, and state tokens.
    These tokens are fused by a GPT-style transformer with \textit{unidirectional attention} and two specialized query sets: \textit{fine-grained affordance} and \textit{action–progress}.
    During training, affordance queries attend to context tokens to predict foresight $\hat{\mathbf{F}}_{t+n}$ with four supervised heads (\texttt{<Global>}, \texttt{<Local>}, \texttt{<Spatial>}, \texttt{<Dynamic>}) that ground future scene understanding.
    At inference, the affordance heads are removed; the action–progress query attends to both context and affordance foresight to condition a diffusion transformer that predicts action $\hat{a}_{t:t+n-1}$ and progress $\hat{p}_{t:t+n-1}$ trajectories for continuous control.
    \textbf{(b) Structured Attention:}
    Affordance subqueries attend only to shared context tokens to stay disentangled, while both query types use causal attention to preserve temporal consistency.
    }
    \label{fig:pipeline}
    \vspace{-0.2cm}
\end{figure*}

\noindent \textbf{Imitation Learning with Progress Supervision.}
Early imitation learning approaches for long-horizon tasks relied on explicit task decomposition, such as symbolic planning \cite{garrett2020pddlstream, jiang2019task} or predefined visual keyframes \cite{chane2021goal, pertsch2020keyframing}. 
Recent works have shifted towards learning continuous progress representations~\cite{bai2025evolve,tan2025robo,zhou2025act2goal,zhai2025vision,lin2025echovla} directly from large-scale video data \cite{nair2022r3m, ma2023liv, majumdar2023we}. These learned progress metrics can be used to reweight behavior cloning \cite{chen2020bail, siegel2020keep,wu2026closed}, assigning higher importance to demonstrations that yield greater progress.
Compared to computationally expensive bootstrapping methods \cite{xu2022discriminator} or strategies that rely on a small, high-quality subset for data labeling \cite{wu2019imitation, wang2023improving}, this intrinsic progress weighting offers improved scalability and stability.
Our approach augments the action space with a scalar progress indicator, used to supervise policy learning for improved long-horizon manipulation consistency.

\noindent \textbf{Affordance Representations for Robotic Manipulation.}
Representing affordances is a core problem in robot learning~\cite{tang2025affordgrasp,tang2025roboafford,hao2025roboafford++,nasiriany2025rt}. Early work generated structured visual outputs from sensor data, predicting affordances as heatmaps over interaction regions \cite{bahl2023affordances, chen2023affordance, geng2022end, ling2024articulated, mo2021where2act, li2023locate, li2024one, yao2026hammer} or as bounding boxes and keypoints to localize relevant parts \cite{huang2024a3vlm, liu2024moka, manuelli2019kpam, qin2020keto, li2025learning}. With VLMs, affordance reasoning became open-vocabulary, either via retrieval from knowledge bases \cite{ju2024robo, kuang2024ram} or via fine-tuning on affordance datasets \cite{qian2024affordancellm, ma2024glover, ma2025glover++}. Recent VLA models further integrate affordance into the policy loop by injecting visual interaction traces \cite{zheng2024tracevla}, predicting affordance plans \cite{nasiriany2025rt}, or encoding affordance types into reasoning chains \cite{li2025coa}. Our approach predicts multiple affordance types as structured latent tokens that provide implicit task-aware representations for progress estimation and subtask policy learning.

\section{Method}

\subsection{Problem Formulation}
\label{sec:prob}
Long-horizon, language-conditioned manipulation requires policies that remain temporally consistent across sub-tasks with heterogeneous dynamics.
Let a robot operate over a distribution of tasks
$\mathcal{T}\!=\!\left\{\tau_k\right\}_{k=1}^K$. Each task $\tau \in \mathcal{T}$ defines an observation–action distribution $p\left({o}_t, {a}_t \mid \tau\right)$ and an implicit temporal phase progression. Differences in background, lighting, object configuration, and viewpoint, compounded by cumulative control errors, alter the observation space. As a result, when sub-policies are trained in isolation, the terminal state of one often deviates from the expected start state of the next, leading to unstable transitions.
Conventional behavior cloning that directly fits $\pi\left({a}_t \mid {o}_t\right)$ on a mixture of demonstrations collapses these distinct phases, causing repetition or skipped subtasks during rollout.
As illustrated in \cref{fig:pipeline}, \modelnamenc addresses long-horizon stability through two tightly coupled components: \textbf{\circnum{1} a fine-grained affordance prediction module} that anticipates future interaction cues by estimating an affordance-centric latent at future offsets \(t+n\), $f_{\mathrm{aff}}\!:\!\mathcal{O}\!\times \mathcal{T}\!\rightarrow\!\mathcal{F}$, serving as a task-relevant intermediate representation that stabilizes perception (\S\ref{sec:aff}), and \textbf{\circnum{2} a progress-aware policy} $\pi\!:\!\mathcal{O}\!\times\!\mathcal{T}\!\rightarrow\!\tilde{\mathcal{A}}$, $\tilde{\mathcal{A}}\!=\!\mathcal{A}\!\times\!\mathcal{P}$, where  $\mathcal{P} \subset[0,1]$ is a continuous progress space. 
At time $t$, given observations ${o}_t\!\in\!\mathcal{O}$, and task specification $\tau\!\in\!\mathcal{T}$, and conditioned on the predicted affordance latent, the policy jointly decodes an action ${a}_t\!\in\!\mathcal{A}$ alongside a scalar $p_t\!\in\!\mathcal{P}$ that encodes progress within the current subtask and serves as a temporal regularizer that mitigates phase ambiguity and enables smooth subtask transitions without relying on separate planners or hierarchical controllers (\S\ref{sec:progress}).

\subsection{\modelname Architecture}
\label{sec:model}
\noindent \textbf{Multi-Modal Encoders.} \modelnamenc processes three synchronized inputs: a language instruction \(l\), an image observation \(o_t\), and a robot state \(s_t\). Instructions are embedded using a CLIP text encoder~\cite{radford2021learning}, observations are encoded with a Masked Autoencoder~\cite{he2022masked} and downsampled by a Perceiver Resampler~\cite{jaegle2021perceiver} to retain task-relevant visual tokens. The robot state is projected through a lightweight MLP. The resulting tokens are concatenated into a single sequence with time index \(t\) to form a unified multimodal sequence passed to the transformer backbone.

\noindent \textbf{Backbone and Learnable Queries.} The backbone follows a GPT-2-style transformer~\cite{radford2019language} to integrate the token sequence with causal and cross-modal attention. 
On top of this backbone, \modelnamenc introduces two learnable query sets that instantiate the affordance and progress-aware policy:

\noindent $\diamondsuit$ \textbf{\textit{Fine-grained affordance queries}} comprise four subqueries \texttt{<Global>}, \texttt{<Local>}, \texttt{<Spatial>}, and \texttt{<Dynamic>}, that attend over language, vision, and state tokens to extract task-relevant representation at complementary scales and produce the affordance-centric latent 
\begin{equation}
\setlength{\abovedisplayskip}{6pt}
\setlength{\belowdisplayskip}{6pt}
\hat{\mathbf{F}}_{t+n}
= f_{\mathrm{aff}}\!\left(l, o_t, s_t\right)
\in \mathcal{F}.
\label{eq:aff-feature}
\end{equation}

\noindent $\diamondsuit$ \textbf{\textit{Action–progress queries}} pool control-relevant context and condition on $\hat{\mathbf{F}}_{t+n}$ to support progress-aware inverse-dynamics decoding for temporally consistent long-horizon control.
Building on prior inverse-dynamics formulations \cite{chen2025villa, tian2024predictive, hu2024video}, these queries aggregate current observations with the predicted affordance latent to infer action sequences that align with the forecasted interaction state. 

\noindent \textbf{Decoders.} The action head is a denoising diffusion transformer~\cite{peebles2023scalable} that conditions on the action–progress queries and the affordance latent to generate a horizon-$n$ trajectory:
\begin{equation}
\setlength{\abovedisplayskip}{6pt}
\setlength{\belowdisplayskip}{6pt}
(\hat{a}_{t:t+n-1}, \hat{p}_{t:t+n-1})\
= \operatorname{DiT}\!\left(
l, o_t, s_t, \hat{\mathbf{F}}_{t+n}
\right),
\label{eq:dit-output}
\end{equation}
where \(\hat{{a}}_{t:t+n-1}\) is the predicted action sequence and \(\hat{p}_{t:t+n-1}\) the predicted sequence of scalar progress values at each step. Details can be found in Appendix~\ref{sec:implement}.

\subsection{Fine-Grained Affordance Prediction}
\label{sec:aff}

\noindent \textbf{Global Prediction.}
Global affordance provides a high-level semantic prior that identifies \emph{which} object in the scene is relevant to the instruction and \emph{where} it is located.
Given a command such as ``pick up the cup,'' the model parses the referent (\ie, the instruction-specified object) at time $t$ and predicts, for a future offset $t\!+\!n$, an object-centric prior that marks the referred instance and its approximate region.
This global prior anchors all subsequent affordance cues and constrains local reasoning to the correct region of interest.

Targets are obtained by first resolving the referent via Grounding DINO~\cite{liu2024grounding} and then segmenting the instance with SAM~\cite{kirillov2023segment} to obtain a binary mask on the observation lattice. An object representation is further extracted by masked pooling over a frozen image encoder within this region. During training, the predicted future mask and object feature are supervised against these targets. The training objective for global affordance tokens is then formulated as:
\begin{equation}
\setlength{\abovedisplayskip}{6pt}
\setlength{\belowdisplayskip}{6pt}
\begin{aligned}
\mathcal{L}_{\text{global}}
&= \mathcal{L}_{\text{FL}}\!\left(
\mathcal{M}_{t+n}^{\text{global}},
\,\hat{\mathcal{M}}_{t+n}^{\text{global}}
\right)
\\
&\quad + \mathcal{L}_{\text{Dice}}\!\left(
\mathcal{M}_{t+n}^{\text{global}},
\,\hat{\mathcal{M}}_{t+n}^{\text{global}}
\right)
\end{aligned}
\end{equation}
where $\hat{\mathcal{M}}_{t+n}^{\text{global}}\!=\!f_{\text{global}}\left(l, o_t, s_t\right)$ denotes the predicted global affordance mask at time $t\!+\!n$, and $\mathcal{M}_{t+n}^{\text {global }}$ is the corresponding binary target mask on the image (observation) lattice. Here, $\mathcal{L}_{\text{FL}}$ denotes the pixel-wise focal loss computed over image domain $\Omega$, and $\mathcal{L}_{\text {Dice }}$ denotes the soft Dice loss. 

\noindent \textbf{Local Prediction.}
Local affordance captures fine-grained geometric localization affordance cues necessary for precise contact reasoning. Conditioned on the region highlighted by the global prior, it analyzes high-frequency local visual structure, such as edges, textures, and part geometry, to predict, at time $t\!+\!n$, a dense contact-likelihood distribution over the object of interest. Following GLOVER++\cite{ma2025glover++}, annotated contact points are converted into Gaussian heatmaps that serve as continuous supervision targets. The model is trained to match this target heatmap distribution.
The training objective for local affordance tokens is thus formulated as:\looseness-1
\begin{equation}
\setlength{\abovedisplayskip}{6pt}
\setlength{\belowdisplayskip}{6pt}
\begin{aligned}
\mathcal{L}_{\text{local}}
&= \mathcal{L}_{\text{FL}}\!\left(
\mathcal{M}_{t+n}^{\text{local}},
\,\hat{\mathcal{M}}_{t+n}^{\text{local}}
\right)
\\
&\quad + \mathcal{L}_{\text{KL}}\!\left(
\tilde{\mathcal{M}}_{t+n}^{\text{local}},
\,\tilde{\hat{\mathcal{M}}}_{t+n}^{\text{local}}
\right),
\label{eq:local-loss}
\end{aligned}
\end{equation}

\begin{equation}
\setlength{\abovedisplayskip}{6pt}
\setlength{\belowdisplayskip}{6pt}
\tilde{\mathcal{M}}
= \frac{\mathcal{M}}
{\sum_{(i,j)\in\Omega} \mathcal{M}^{(i,j)} + \varepsilon},
\label{eq:normalize}
\end{equation}
where $\hat{\mathcal{M}}_{t+n}^{\text {local }}\!=\!f_{\text {local }}\left(l, o_t, s_t\right)$ is the predicted future local contact-likelihood map at $t\!+\!n, \mathcal{M}_{t+n}^{\text {local }}$ is the soft Gaussian target heatmap, and $\tilde{\mathcal{M}}$ denotes the $\ell_1$-normalized map used in the KL term with a small $\varepsilon>0$.

\noindent \textbf{Spatial Prediction.}
Spatial affordance converts underspecified spatial language into executable placement proposals that remain robust under layout changes.
Instead of memorizing a single coordinate, the model predicts a small set of candidate placement points at time  \(t\!+\!n\), each representing a plausible region for placing the object.
Ground-truth targets are obtained by first converting the instruction into spatial semantics using SpatialVLM~\cite{chen2024spatialvlm} and then sampling executable 2D coordinates with RoboPoint~\cite{yuan2024robopoint}.
The training objective minimizes a set-matching objective that aligns each target point with its closest predicted candidate:
\begin{equation}
\setlength{\abovedisplayskip}{6pt}
\setlength{\belowdisplayskip}{6pt}
\mathcal{L}_{\text{spatial}}
= 
\frac{1}{C_{t+n}} \sum_{c=1}^{C_{t+n}}
\min_{1 \leq m \leq M}
\left\|
\hat{\mathbf{p}}_{t+n}^{(m)} - \mathbf{p}_{t+n}^{(c)}
\right\|_2^2
\end{equation}
where $\hat{\mathcal{S}}_{t+n}\!=\!\{\hat{\mathbf{p}}_{t+n}^{(m)}\}_{m=1}^M$ and $\mathcal{S}_{t+n}\!=\!\{{\mathbf{p}}_{t+n}^{(c)}\}_{c=1}^{C_{t+n}}$ are the set of $M$ predicted normalized 2D candidates and $C_{t+n}$ target placement points at $t\!+\!n$, respectively.\looseness-1

\noindent \textbf{Dynamic Prediction.}
Dynamic affordance identifies the pixels corresponding to the robot gripper and movable objects, and predicts how these regions will evolve over time. Its goal is to predict and establish the statistical associations among the current scene, the language instruction, and the actions required to realize the predicted motion. 
To construct supervision, we apply a grid-based tracking protocol: an \(N\!\times\!N\) visual grid of query points is placed on the first frame of a short history at \(t-\delta\). CoTracker~\cite{karaev2024cotracker} follows each point forward in time. We compute cumulative displacement for each trajectory and retain those that exceed a threshold, ensuring that the dynamic mask captures genuine motion rather than static background or tracker jitter. The selected trajectories are then rasterized at \(t\!+\!n\) into a dynamic region. Given inputs at time \(t\), the model predicts a future dynamic probability map that highlights pixels likely to belong to the gripper or other moving objects at \(t\!+\!n\). Training encourages calibrated per-pixel probabilities and alignment with the tracked motion regions. The training objective for dynamic affordance tokens is formulated as a masked reconstruction loss using a latent-variable model:
\begin{equation}
\setlength{\abovedisplayskip}{6pt}
\setlength{\belowdisplayskip}{6pt}
\begin{split}
\mathcal{L}_{\text{dynamic}}
=
\mathbb{E}_{\mathbf{z}\sim Q_\phi\!\left(\mathbf{z}\mid x_{t+n}^{\mathcal{M}}\right)}
\!\left[-\log P_\psi\!\left(x_{t+n}^{\mathcal{M}} \mid \mathbf{z}\right)\right]
\\
\qquad
+ \beta \operatorname{KL}\!\Big(Q_\phi\!\left(\mathbf{z} \mid x_{t+n}^{\mathcal{M}}\right) \,\|\, p(\mathbf{z})\Big)
\end{split}
\end{equation}
\noindent where 
\(\mathcal{M}_{t+n}\) is the dynamic region mask from tracking, \(x_{t+n}^{\mathcal{M}}\) is the future frame restricted to those masked pixels, \(Q_\phi(\mathbf{z}\mid x_{t+n}^{\mathcal{M}})\) is the posterior, \(P_\psi(x_{t+n}^{\mathcal{M}}\mid\mathbf{z})\) the decoder likelihood, \(p(\mathbf{z})\) the prior, and \(\beta\) the KL weight.

\subsection{Progress-aware Policy via Inverse Dynamics}
\label{sec:progress}
In addition to predicting where to act via affordances, we introduce a progress-aware prediction task that estimates how far execution has advanced within the current subtask. At each time step, \modelnamenc first infers the active subtask stage from the affordance latent and derives a stage embedding. Conditioned on this embedding, the model predicts a scalar $p_t \in[0,1]$ that quantifies within-stage completion.
We append this scalar to the action output so the policy jointly predicts
$\left(a_t, p_t\right)$ under a shared multimodal context. This explicit progress signal reduces ambiguity in long-horizon control: visually similar observations may correspond to different actions depending on stage, and $p_t$ disambiguates these cases by providing a continuous indicator of ``where we are'', stabilizing learning and execution by encouraging monotonic, stage-consistent evolution of the latent state and by smoothing transitions at sub-policy boundaries without relying on separate high-level controllers.

Classical inverse dynamics predicts the action $\hat{a}_t$ bridging two temporally ordered observations $(o_t, o_{t+1})$. We extend this to predict an $n$-step action–progress sequence conditioned on the current inputs and a single-step affordance latent. We instantiate $f_{\mathrm{inv}}$ as a denoising diffusion transformer that conditions on the current observation $o_t$, the instruction $l$, the robot state $s_t$, and the predicted affordance latent $\hat{\mathbf{F}}_{t+n}\!=\!f_{\mathrm{aff}}(l,o_t,s_t)$ to generate
\begin{equation}
\setlength{\abovedisplayskip}{6pt}
\setlength{\belowdisplayskip}{6pt}
\left(\hat{a}_{t:t+n-1},\, \hat{p}_{t:t+n-1}\right)
= f_{\mathrm{inv}}\!\left(l, o_t, s_t, \hat{\mathbf{F}}_{t+n}\right).
\label{eq:inverse-model}
\end{equation}
The DiT head jointly models action distributions and progress values, capturing correlations among these cross-modal inputs across the trajectory. Training follows the standard diffusion objective:
\begin{equation}
\setlength{\abovedisplayskip}{6pt}
\setlength{\belowdisplayskip}{6pt}
\tilde{\mathbf{y}}_{t:t+n-1,t_d}
= \sqrt{\bar{\alpha}_{t_d}}\,\mathbf{y}_{t:t+n-1}
+ \sqrt{1-\bar{\alpha}_{t_d}}\,\boldsymbol{\epsilon}
\label{eq:yt-perturb}
\end{equation}
\begin{equation}
\mathcal{L}_{\text{DiT}}=\mathbb{E}_{t_d,\,\boldsymbol{\epsilon}}
\big\|
\boldsymbol{\epsilon}
-
\epsilon_{\theta}\!\left(
\tilde{\mathbf{y}}_{t:t+n-1,t_d}\,\middle|\,l, o_t, s_t, \hat{\mathbf{F}}_{t+n}, t_d
\right)
\big\|_2^{2}
\label{eq:diff-loss-packed}
\end{equation}
where $\mathbf{y}_{t:t+n-1}$ is the target action–progress vector over steps $t$ to $t{+}n{-}1$, $\boldsymbol{\epsilon} \sim \mathcal{N}(\mathbf{0}, \mathbf{I})$ is Gaussian noise, $t_d$ is the diffusion time with $\bar{\alpha}_{t_d}$ the cumulative noise schedule, $\tilde{\mathbf{y}}_{t:t+n-1, t_d}$ is the noised target, and $\epsilon_\theta(\cdot)$ is the noise predictor.

\begin{table*}[t]
\centering
\caption{\textbf{CALVIN ABC$\rightarrow$D experimental results.} We group the baselines into four types and report the average success rate of the top three checkpoints, computed over 1,000 rollouts per task, as well as the average number of consecutively completed tasks to solve 5 instructions (Avg. Len.). PALM consistently and substantially outperforms all baselines. Best results are shown in \textbf{bold}.}
\renewcommand{\arraystretch}{1}
\setlength\tabcolsep{15pt}
\vspace{-0.3cm}
\resizebox{\linewidth}{!}{
\begin{tabular}{lcccccccc}
\toprule
\multirow{2}{*}{\textbf{Method}} &
\multirow{2}{*}{\textbf{Type}} &
\multicolumn{6}{c}{\textbf{Task completed in a row}} \\ \cmidrule{3-8}
 & & \textbf{1} & \textbf{2} & \textbf{3} & \textbf{4} & \textbf{5} & \textbf{Avg. Len.} $\uparrow$ \\ 
\midrule
RT-1~\cite{brohan2022rt} & Autoregressive & 53.3\% & 22.2\% & 9.40\% & 3.80\% & 1.30\% & 0.90 \\
Robo-Flamingo~\cite{li2023vision} & Autoregressive & 82.4\% & 61.9\% & 46.6\% & 33.1\% & 23.5\% & 2.47 \\ 
OpenVLA~\cite{kim2024openvla} & Autoregressive & 91.3\% & 77.8\% & 62.0\% & 52.1\% & 43.5\% & 3.27\\ 
Diffusion Policy~\cite{chi2025diffusion} & Diffusion-based & 40.2\% & 12.3\% & 2.60\% & 0.80\% & 0.00\% & 0.56 \\
$\pi_0$~\cite{black2024pi0visionlanguageactionflowmodel} & Diffusion-based & 93.8\% & 85.0\% & 76.7\% & 68.1\% & 59.9\% & 3.92 \\
3D-VLA~\cite{zhen20243d} & 3D-Aware & 44.7\% & 16.3\% & 8.10\% & 1.60\% & 0.00\% & 0.71 \\ 
3D Diffuser Actor~\cite{ke20243d} & 3D-Aware & 92.2\% & 78.7\% & 63.9\% & 51.2\% & 41.2\% & 3.27 \\ 
RoboUniview~\cite{liu2024robouniview} & 3D-Aware & 94.2\% & 84.2\% & 73.4\% & 62.2\% & 50.7\% & 3.65 \\ 
Susie~\cite{black2023zero} & Prediction & 87.0\% & 69.0\% & 49.0\% & 38.0\% & 26.0\% & 2.69 \\
GR-1~\cite{wu2023unleashing} & Prediction & 85.4\% & 71.2\% & 59.6\% & 49.7\% & 40.1\% & 3.06 \\ 
Seer~\cite{tian2024predictive} & Prediction & 94.4\% & 87.2\% & 79.9\% & 72.2\% & 64.3\% & 3.98 \\ 
\midrule
\rowcolor{gray!20} \modelname~(\xmark~progress) & Prediction & 95.3\% & 85.6\% & 79.5\% & 74.3\% & 67.0\% & 4.02 \\
\rowcolor{gray!20}
\textbf{\modelname} & \textbf{Prediction + Progress } & 
\textbf{96.9\%} & \textbf{93.8\%} & \textbf{89.3\%} & \textbf{85.9\%} & \textbf{82.0\%} & \textbf{4.48} \\
\bottomrule
\end{tabular}
}
\label{table1}
\vspace{-0.1cm}
\end{table*}

\section{Experiments}

\noindent \textbf{Training Details.}
Our training process consists of a pre-training and a fine-tuning stage. For pre-training, we utilize a mixed dataset from the DROID~\cite{khazatsky2024droid} and BridgeData V2~\cite{walke2023bridgedata} datasets, which together provide large-scale, in-the-wild robotic arm demonstrations to build a foundational understanding of diverse real-world tasks, and EPIC-KITCHENS~\cite{damen2020epic} and RoboCerebra~\cite{han2025robocerebra}, which provide fine-grained sub-steps and time-segment annotations to supervise the learning of semantic progress estimation in long-horizon, contact-rich scenarios. For fine-tuning, we select 942 trajectories from robot data and annotate them with affordance data and continuous progress labels using a semi-automated method. Details can be found in Appendix~\ref{sec:implement}.

\subsection{Simulation Experiments}
\noindent\textbf{Benchmarks.} We conduct evaluations across two simulation benchmarks: the \textbf{LIBERO}~\cite{liu2023libero} benchmark comprising
four distinct task suites (Spatial, Object, Goal, Long), each with 10 tasks and 50 demonstrations for evaluating the robot's comprehension of spatial relationships, and the \textbf{CALVIN}~\cite{mees2022calvin} benchmark, designed to evaluate the instruction-following capabilities of robotic policies on long-horizon language-conditioned tasks, comprising 34 tasks across four distinct environments. We focus on the challenging ABC$\rightarrow$D setting, where we pre-train in the ABC environments and evaluate in the unseen D environment.
\begin{table*}[t!]
\centering
\caption{\textbf{LIBERO experimental results.} 
For each task suite (Spatial, Object, Goal, Long), we report the average success rate and standard error across $3$ seeds with $500$ episodes each. PALM achieves the best performance over previous methods. Best results in \textbf{bold}.}
\vspace{-0.3cm}
\setlength\tabcolsep{20pt}
\resizebox{\linewidth}{!}{
\begin{tabular}{lccccc}
\toprule
 & Average ($\uparrow$) & Spatial ($\uparrow$) & Object ($\uparrow$) & Goal ($\uparrow$) & Long ($\uparrow$) \\
\midrule
OpenVLA~\cite{kim2024openvla} & 76.5 $\pm$ 0.6\% & 84.7 $\pm$ 0.9\% & 88.4 $\pm$ 0.8\% & 79.2 $\pm$ 1.0\% & 53.7 $\pm$ 1.3\% \\
Diffusion Policy~\cite{chi2025diffusion} & 72.4 $\pm$ 0.7\% & 78.3 $\pm$ 1.1\% & 92.5 $\pm$ 0.7\% & 68.3 $\pm$ 1.2\% & 50.5 $\pm$ 1.3\% \\
Octo fine-tuned~\cite{team2024octo} & 75.1 $\pm$ 0.6\% & 78.9 $\pm$ 1.0\% & 85.7 $\pm$ 0.9\% & 84.6 $\pm$ 0.9\% & 51.1 $\pm$ 1.3\% \\
SpatialVLA~\cite{qu2025spatialvla} & 69.0 $\pm$ 1.2\% & 88.2 $\pm$ 0.7\% & 89.9 $\pm$ 1.3\% & 78.6 $\pm$ 0.9\% & 55.5 $\pm$ 1.5\% \\
CoT-VLA~\cite{zhao2025cot} & 81.1 $\pm$ 0.6 \%& 87.5 $\pm$ 1.4\% & 91.6 $\pm$ 0.5\% & 87.6 $\pm$ 0.6\% & 69.0 $\pm$ 0.8\%\\
TraceVLA~\cite{zheng2024tracevla} & 74.8 $\pm$ 0.9\% & 84.6 $\pm$ 1.0\% & 85.2 $\pm$ 0.6\% & 75.1 $\pm$ 1.4\% & 54.1 $\pm$ 1.0\% \\
CoA-VLA~\cite{li2025coa} & 79.8 $\pm$ 0.5\% & 85.3 $\pm$ 0.9\% & 93.1 $\pm$ 1.0\% & 85.8 $\pm$ 0.9\% & 55.0 $\pm$ 1.2\% \\
\midrule
\rowcolor{gray!20} \modelname & \textbf{94.5 $\pm$ 1.0\%} & \textbf{95.2 $\pm$ 1.2 \%}& \textbf{96.7 $\pm$ 0.7\%} & \textbf{94.3 $\pm$ 1.6\%} & \textbf{91.8 $\pm$ 0.8\%} \\
\bottomrule
\end{tabular}
}
\label{tab:libero_performance}
\end{table*}

\noindent\textbf{Baselines.} We compare \modelnamenc with four types of baselines: autoregressive methods, diffusion-based methods, 3D-aware methods and prediction methods. For the CALVIN ABC$\rightarrow$D benchmark, we compare to RT-1~\cite{brohan2022rt}, Robo-Flamingo~\cite{li2023vision}, and OpenVLA~\cite{kim2024openvla}, which are autoregressive action models that generate actions from pre-trained VLMs. Diffusion Policy~\cite{chi2025diffusion} and $\pi_0$~\cite{black2024pi0visionlanguageactionflowmodel} are selected as representative diffusion-based methods that use a denoising diffusion process to model high-dimensional action distributions, while 3D-VLA~\cite{zhen20243d}, 3D Diffuser Actor~\cite{ke20243d}, and RoboUniview~\cite{liu2024robouniview} specialize in capturing 3D-aware representations to enhance manipulation. Prediction-based methods, represented by Susie~\cite{black2023zero}, GR-1~\cite{wu2023unleashing}, and Seer~\cite{tian2024predictive}, merge visual foresight as a future representation to enhance performance in multi-task robot manipulation. For LIBERO, we compare with OpenVLA, Diffusion Policy, and Seer. Additionally, we compare against Octo~\cite{team2024octo}, which pre-trains robot policies on diverse datasets to enhance generalization. SpatialVLA~\cite{qu2025spatialvla}, CoT-VLA~\cite{zhao2025cot}, TraceVLA~\cite{zheng2024tracevla}, and CoA-VLA~\cite{li2025coa} are included to compare diverse task representation methods.\looseness-1

\noindent \textbf{Results.}  \Cref{table1} summarizes performance on the CALVIN ABC$\rightarrow$D long-horizon benchmark, demonstrating that \modelnamenc achieves state-of-the-art results across all metrics and outperforms all baselines. First, \modelnamenc (Prediction + Progress) reaches a 96.9\% success rate on the first subtask and maintains strong performance as horizon length increases (\eg, 82.0\% for five consecutive subtasks). This is a +17.7\% absolute improvement over the strongest prior baseline (Seer at 64.3\%)
at the 5-task horizon. \modelnamenc also yields the longest average task trajectory (4.48), exceeding Seer (3.98) and $\pi_0$ (3.92), indicating significantly more stable execution over extended sequences.
Importantly, results confirm that the progress-aware policy is critical for long-horizon generalization, as removing the progress prediction (\modelnamenc \xmark ~progress) reduces performance consistently across horizons (\eg, average length 4.02 $\rightarrow$ 4.48). 

Moreover, as shown in \Cref{tab:libero_performance}, across all four LIBERO suites, \modelnamenc achieves state-of-the-art performance with an average success rate of 94.5\%. The largest gain is in LIBERO-LONG, where \modelnamenc reaches 91.8\%, outperforming the strongest baseline (CoT-VLA at 69.0\%) by 22.8\%.\looseness-1

\subsection{Ablation Studies}

\begin{figure}[t!]
    \centering
    \includegraphics[width=0.99\linewidth]{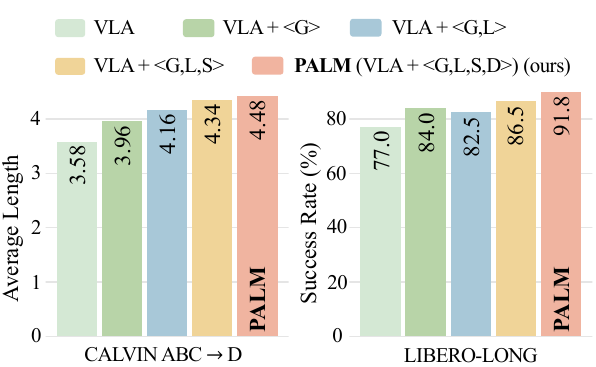}
    \vspace{-0.3cm}
    \caption{\textbf{Ablation studies of affordance components} 
    on CALVIN ABC$\rightarrow$D and LIBERO-LONG benchmarks demonstrate the effectiveness of the four components of affordance prediction.}
    \label{fig:ablation_q1}
    \vspace{-0.2cm}
\end{figure}

\begin{table}[t!]
\captionof{table}{\textbf{Ablation studies of \modelnamenc{} components.} Results on the CALVIN ABC$\rightarrow$D benchmark demonstrate the effectiveness of each training module under both pre-training and fine-tuning.}
\vspace{-0.3cm}
\begingroup
\resizebox{\columnwidth}{!}{%
\begin{tabular}{ccc}
\toprule
\multirow{2}{*}{Ablation Type} & 
Pre-training & 
Fine-tuning  \\ \cmidrule(lr){2-2} \cmidrule(lr){3-3}
& Avg. Len. $\uparrow$ & Avg. Len. $\uparrow$  \\
\midrule
\rowcolor{gray!20} \textbf{\modelname} & \textbf{4.48} & \textbf{4.48} \\ 
\xmark~Affordance Foresight & 3.90  & 3.58 \\ 
\xmark~Inverse Dynamic Prediction & 4.17 & 3.92  \\ 
\xmark~Progress Prediction & 3.73 & 4.02  \\ 
\bottomrule
\end{tabular}%
}
\label{tab:arch}
\endgroup
\vspace{-0.2cm}
\end{table}

\noindent \textbf{How do the components of the fine-grained affordance module affect performance?}
We evaluate the four types of predictable affordance latents: \texttt{Global}, \texttt{Local}, \texttt{Spatial}, and \texttt{Dynamic} to measure their respective contributions. We train models by cumulatively adding these affordances to a vanilla VLA. As illustrated in \Cref{fig:ablation_q1}, adding Global affordance (G) already yields a consistent improvement on both CALVIN and LIBERO-LONG, indicating that coarse object-centric cues help stabilize long-horizon reasoning. Incorporating Local affordance (L) provides additional gains on CALVIN but introduces a slight dip on LIBERO-LONG, likely due to viewpoint-induced geometric bias that affects fine-grained edge-based features. Adding Spatial affordance (S) restores and further improves performance across both benchmarks by providing robust placement priors that generalize across layout variations. Finally, adding Dynamic affordance (D) to the full set (\modelnamenc{}) yields the best performance, indicating that combining motion cues with structured spatial reasoning produces the most reliable long-horizon behavior.\looseness-1

\noindent \textbf{How do the proposed \modelname modules affect performance?}
We study the contributions of \modelnamenc{}'s three core components (affordance prediction module, inverse dynamics prediction, and progress-aware prediction) by conducting ablations during both pre-training and fine-tuning on the CALVIN ABC$\rightarrow$D benchmark. 
Table~\ref{tab:arch} shows that all three \modelnamenc{} components contribute complementary capabilities and collectively result in the best performance. Removing affordance foresight produces the largest drop in fine-tuning performance (4.48 $\rightarrow$ 3.58), confirming that structured future affordance prediction is essential for precise long-horizon planning once the policy is adapted to downstream robot data. In contrast, removing progress prediction causes the largest degradation during pre-training (4.48 $\rightarrow$ 3.73), highlighting that large-scale long-horizon datasets are particularly valuable for learning a strong progress prior that stabilizes temporal reasoning. Eliminating inverse dynamics prediction also reduces performance in both stages, showing that predicting multi-step action trajectories conditioned on affordance latents provides an important training signal beyond direct behavior cloning.
\begin{table}[t!]
\begingroup
\captionof{table}{\textbf{Ablation studies on training data composition.} Results on the CALVIN ABC$\rightarrow$D and LIBERO-LONG benchmarks demonstrate the data efficiency of each source type.}
\vspace{-0.3cm}
\resizebox{\columnwidth}{!}{
\begin{tabular}{ccc}
\toprule
\multirow{2}{*}{Ablation Type} & 
CALVIN ABC$\rightarrow$D & 
LIBERO-LONG  \\ \cmidrule(lr){2-2} \cmidrule(lr){3-3}
& Avg. Len. $\uparrow$ & SR (\%) $\uparrow$  \\
\midrule
\rowcolor{gray!20} \textbf{\modelname} & \textbf{4.48} & \textbf{91.8} \\ 
\xmark~In-the-Wild Data & 3.90  & 73.5 \\ 
\xmark~Long-Horizon Video Data & 3.73 & 84.5  \\ 
\xmark~Human Annotated Data & 3.58 & 76.5  \\ 
\xmark~Simulation Data (Pretrain)  & 3.96 & 81.0  \\ 
\bottomrule
\end{tabular}%
}  
\label{tab:data}
\endgroup
\vspace{-0.2cm}
\end{table}

\begin{figure*}[t!]
    \centering
\includegraphics[width=0.9\linewidth]{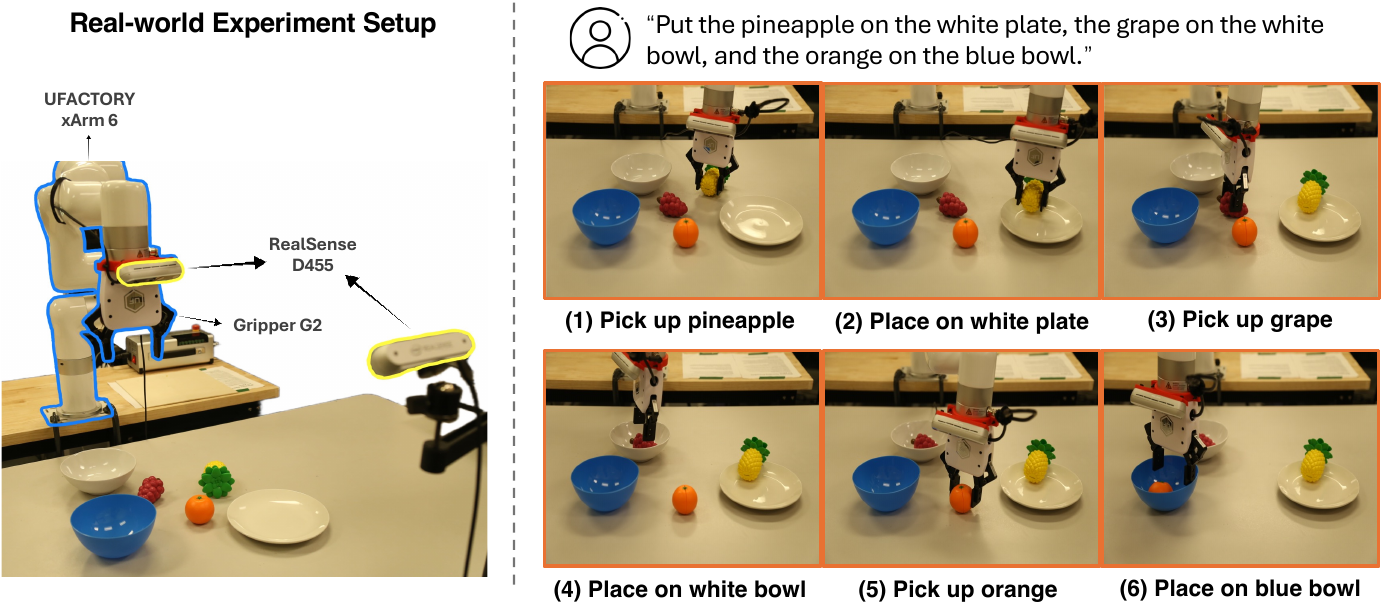}
    \vspace{-0.3cm}
    \caption{\textbf{Real-world experimental setup and task design.} \textbf{Left:} We use a UFACTORY xArm6 robot with the matched Gripper G2 and two RealSense D455 cameras. \textbf{Right:} We design a real-world long-horizon manipulation task consisting of six consecutive subtasks, driven by a single high-level instruction.}
    \label{fig:real}
    \vspace{-0.2cm}
\end{figure*}

\begin{table*}[t!]
\centering
\caption{\textbf{Real-world experimental results} on long-horizon task under different generalization settings.}
\vspace{-0.3cm}
\setlength\tabcolsep{15pt}
\begin{tabular}{@{}c@{}}
\resizebox{0.93\linewidth}{!}{%
\begin{tabular}{c c *{6}{c} c}
\toprule
\multirow{2}{*}{\textbf{Type}} & \multirow{2}{*}{\textbf{Method}}
  & \multicolumn{6}{c}{\textbf{Task completed in a row}} & \multirow{2}{*}{\textbf{Avg. Len.} $\uparrow$} \\
\cmidrule(lr){3-8}
& & \textbf{1} & \textbf{2} & \textbf{3} & \textbf{4} & \textbf{5} & \textbf{6} & \\
\midrule
\multirow{3}{*}{Random Localization}
& OpenVLA~\cite{kim2024openvla} & 0.45 & 0.30 & 0.15 & 0.05 & 0.00 & 0.00 & 0.95 \\
& Octo~\cite{team2024octo}    & 0.35 & 0.20 & 0.10 & 0.00 & 0.00 & 0.00 & 0.65 \\
& \cellcolor{gray!20}\textbf{\modelname} &
\cellcolor{gray!20}\textbf{0.70} & \cellcolor{gray!20}\textbf{0.65} &
\cellcolor{gray!20}\textbf{0.55} & \cellcolor{gray!20}\textbf{0.45} &
\cellcolor{gray!20}\textbf{0.40} & \cellcolor{gray!20}\textbf{0.30} &
\cellcolor{gray!20}\textbf{3.05} \\
\midrule
\multirow{3}{*}{Visual Distraction}
& OpenVLA~\cite{kim2024openvla} & 0.65 & 0.50 & 0.25 & 0.15 & 0.05 & 0.00 & 1.60 \\
& Octo~\cite{team2024octo}    & 0.45 & 0.35 & 0.15 & 0.00 & 0.00 & 0.00 & 0.95 \\
& \cellcolor{gray!20}\textbf{\modelname} &
\cellcolor{gray!20}\textbf{0.85} & \cellcolor{gray!20}\textbf{0.80} &
\cellcolor{gray!20}\textbf{0.65} & \cellcolor{gray!20}\textbf{0.60} &
\cellcolor{gray!20}\textbf{0.50} & \cellcolor{gray!20}\textbf{0.40} &
\cellcolor{gray!20}\textbf{3.80} \\
\midrule
\multirow{3}{*}{Unseen Lighting}
& OpenVLA~\cite{kim2024openvla} & 0.55 & 0.35 & 0.25 & 0.10 & 0.00 & 0.00 & 1.25 \\
& Octo~\cite{team2024octo}    & 0.50 & 0.35 & 0.15 & 0.05 & 0.00 & 0.00 & 1.05 \\
& \cellcolor{gray!20}\textbf{\modelname} &
\cellcolor{gray!20}\textbf{0.80} & \cellcolor{gray!20}\textbf{0.70} &
\cellcolor{gray!20}\textbf{0.60} & \cellcolor{gray!20}\textbf{0.60} &
\cellcolor{gray!20}\textbf{0.45} & \cellcolor{gray!20}\textbf{0.40} &
\cellcolor{gray!20}\textbf{3.55} \\
\bottomrule
\end{tabular}
}%
\end{tabular}
\label{tab:long_horizon}
\end{table*}

\noindent \textbf{Effectiveness of training data composition for robot manipulation.} Training data is a critical factor influencing the performance of robot policies. Accordingly, we classify our training data into four primary types based on their task orientation: (1) In-the-wild datasets, represented by  DROID~\cite{khazatsky2024droid} and BridgeData V2~\cite{walke2023bridgedata} datasets for the pre-training stage, (2) Long-horizon video datasets, consisting of EPIC-KITCHENS~\cite{damen2020epic} and RoboCerebra~\cite{han2025robocerebra}, which are also utilized during pre-training to learn the understanding of semantic progress in complex, multi-step scenarios, (3) Human-annotated datasets, employed in the fine-tuning stage to train affordance and progress-aware policy, and (4) Simulation datasets for post-training evaluation (CALVIN ABC$\rightarrow$D~\cite{mees2022calvin} and LIBERO-LONG~\cite{liu2023libero}).

As illustrated in \Cref{tab:data}, excluding any single data type invariably degrades model performance, though to varying extents. 
This degradation is most pronounced for the In-the-Wild and Human-Annotated datasets. A substantial performance decrease also occurs when removing the pre-training data corresponding to the simulation benchmarks.
Additional ablation studies are provided in Appendix~\ref{sec:extra_exp}.

\subsection{Real-World Experiments}
\noindent \textbf{Experimental Setup.} As shown in Figure~\ref{fig:real}, we use a UFACTORY xArm6 with a Gripper G2 to conduct real-world experiments. For visual input, we use two RealSense D455 cameras configured as eye-on-hand and eye-on-base to capture RGB images. We design several types of generalization tests for long-horizon tasks, including Visual Distraction, Random Localization, and Unseen Lighting. 
We select a mixed pre-training dataset composed of DROID~\cite{khazatsky2024droid} and BridgeData V2~\cite{walke2023bridgedata}, while the fine-tuning dataset consists of 200 demonstrations collected on the xArm with RGB images, robot states, and actions.

\noindent \textbf{Baselines and Metrics.} We select OpenVLA~\cite{kim2024openvla} and Octo~\cite{team2024octo} as baselines. We report the success rate (SR) and average length for each task over 20 real-world rollouts. For each rollout, each method is permitted a maximum of three execution attempts. To ensure fairness, all models are fine-tuned on our training dataset, trained for an equal number of iterations, and evaluated with the final checkpoint.

\noindent \textbf{Generalization Evaluation for Long-Horizon Task.}
To evaluate generalization in long-horizon tasks, we construct a sequential pick-and-place task consisting of 6 consecutive subtasks, driven by a single high-level instruction. We consider three generalization settings: \circnum{1} varying the target object pose, \circnum{2} changing the scene lighting to unseen conditions, and \circnum{3} adding multiple distractor objects to induce visual clutter. As shown in Table~\ref{tab:long_horizon}, results demonstrate \modelnamenc{}’s superior generalization over baselines as the task sequence length increases, showing its robustness in long-horizon settings. Details are available in Appendix~\ref{sec:realworld}.

\section{Conclusion}
We introduce \modelname, a vision–language–action (VLA) model for long-horizon robotic manipulation that couples structured future affordance prediction with continuous progress estimation in a closed loop. \modelnamenc{} achieves state-of-the-art results on two benchmarks, with a 12.5\% improvement on CALVIN ABC$\rightarrow$D and 91.8\% success on LIBERO-LONG, and shows significant robustness in real-world experiments across long-horizon generalization settings.

{
    \small
    \bibliographystyle{ieeenat_fullname}
    \bibliography{main}
}


\clearpage
\appendix
\clearpage
\setcounter{page}{1}
\maketitlesupplementary

\section{Implementation Details}
\label{sec:implement}

\subsection{\modelname{} Model Details}
\label{sec:model_detail}
\noindent \textbf{Vision.} Visual encoding employs a MAE-pretrained ViT-B~\cite{he2022masked} backbone, which serves as the primary feature extractor. For each timestep, the model processes images from two viewpoints: a static eye-on-base camera for global context and an eye-on-hand camera for local, gripper-centric views. The ViT transforms each image into 196 patch embeddings plus a [CLS] token. To maintain computational tractability over long temporal sequences, we employ a Perceiver Resampler~\cite{jaegle2021perceiver}. This module uses a set of learnable latent vectors and cross-attention to distill the initial 197 high-dimensional tokens into a compact, fixed-size set of task-relevant visual embeddings, which are then fed to the main backbone.\looseness-1

\noindent \textbf{Text.} To ground the policy in natural language, we encode task instructions using a pretrained CLIP text encoder~\cite{radford2021learning}. This module converts the input instruction into a fixed-length embedding that captures its semantic intent. A subsequent linear projection maps this embedding into the model’s shared latent space, allowing effective integration with visual and state information.

\noindent \textbf{Robot State.} The model receives proprioceptive feedback describing the robot’s physical configuration. This state is represented by a six-dimensional vector for the end-effector’s 6-DoF Cartesian pose (position and Euler angles) and a binary value for the gripper’s open/closed status. For processing, the binary gripper value is first transformed into a two-dimensional one-hot vector. Both the 6-D pose and the one-hot gripper vector are projected through separate linear layers, then concatenated and passed through a final MLP to generate a single, unified state token.

\noindent \textbf{Learnable Queries.} We introduce two sets of learnable query tokens that are appended to the multimodal token sequence at each step $t$ and updated inside the transformer via cross-attention under causal masking. Each query set extracts and integrates information from multimodal inputs to enable joint prediction.
\begin{itemize}
\item \textbf{Fine-grained affordance queries} are specialized tokens that extract a structured view of future interaction. Organized into four sub-queries (\texttt{<Global>}, \texttt{<Local>}, \texttt{<Spatial>}, \texttt{<Dynamic>}), they attend to the multimodal context to produce a disentangled, affordance-centric latent $\hat{\mathbf{F}}_{t+n}$ that guides downstream control.
\item \textbf{Action–progress queries} generate the final control sequence. They pool control-relevant information from current observations and explicitly condition on $\hat{\mathbf{F}}_{t+n}$, enabling a progress-aware inverse-dynamics formulation so actions remain temporally consistent and aligned with the predicted future state.
\end{itemize}

\noindent \textbf{Backbone.} The core of our model is a GPT-2 style transformer~\cite{radford2019language}, which functions as the central multi-modal reasoning engine. It takes as input a concatenated sequence containing the encoded vision, text, and state tokens, alongside the learnable affordance and action-progress queries. By applying causal and cross-modal attention, the backbone fuses these diverse inputs into a coherent latent representation of the world state.

\noindent \textbf{Decoders.} We preserve spatial correspondence by adding fixed 2D sine–cosine positional encodings to the image patch tokens and propagating these coordinates through a stack of Transformer encoder layers. After this shared encoding, modality-specific heads decode the four affordance latents. \texttt{<Global>} and \texttt{<Spatial>} use lightweight 2-layer MLP heads, producing a future object mask and a set of candidate placement points. \texttt{<Local>} and \texttt{<Dynamic>} use 2-layer Transformer blocks followed by linear projections to produce a future contact heatmap and a dynamic region. These targets are defined at $t+n$ to supervise the affordance latents and are used only during training. The primary model output is produced by the action–progress decoder, which jointly predicts actions and scalar progress. Configuration details for each module are given in Table~\ref{tab:module_config}.

\begin{table*}[t!]
    \centering
    \setlength{\tabcolsep}{7pt}
    \caption{\textbf{Key parameters of each module in \modelnamenc{}.}}
    \label{tab:module_config}
    \vspace{-0.3cm}
    \begin{tabular}{lccc}
        \toprule
        \textbf{Type} & \textbf{Hidden Size} & \textbf{Number of Layers} & \textbf{Number of Heads} \\
        \midrule
        Image Encoder     & 768  & 12 & 12 \\
        Perceiver Resampler & 768  & 3  & 8  \\
        GPT-2 (LLM Backbone)   & 384 & 24 & 12 \\
        Global Decoder    & 384 & 2  & 16 \\
        Local Decoder     & 384 & 2  & 16 \\
        Spatial Decoder   & 384 & 2  & 16 \\
        Dynamic Decoder   & 384 & 2  & 16 \\
        \bottomrule
    \end{tabular}
\end{table*}

\noindent \textbf{Prediction with Diffusion Transformer.} We formulate action–progress generation as a conditional denoising task and employ a Diffusion Transformer (DiT) decoder~\cite{peebles2023scalable}. The decoder conditions on the latent embedding from the action–progress queries, which already integrates the predicted affordance information. By iteratively reversing a Gaussian noise process over a sequence of latent vectors, the DiT models complex, multimodal action distributions and yields a temporally coherent trajectory. It produces a joint sequence of 7-DoF actions and the corresponding scalar progress values associated with subtask completion. Configuration details of the DiT are provided in Table~\ref{tab:dit_config}.

\begin{table}[h]
\centering
\caption{\textbf{Configuration of the Diffusion Transformer} used for action-progress prediction.}
\label{tab:dit_config}
\vspace{-0.3cm}
\begin{tabular}{ll}
\toprule
\textbf{Parameter} & \textbf{Value} \\
\midrule
Hidden Size & 384 \\
Number of Layers & 12 \\
Number of Heads & 12 \\
Sampling Steps & 10 \\
Noise Schedule & Cosine \\
Action Prediction Steps & 3 \\
Loss Function & MSE ($L_2$ loss) \\
Precision & fp32 \\
\bottomrule
\end{tabular}
\vspace{-0.4cm}
\end{table}

\subsection{Training Details}
\noindent \textbf{Datasets.} Our training process consists of a pre-training and a fine-tuning stage. For pre-training, we utilize a mixed dataset from two domains. One part is drawn from the DROID~\cite{khazatsky2024droid} dataset and BridgeData V2~\cite{walke2023bridgedata}, which together provide large-scale, in-the-wild robotic arm demonstrations to build a foundational understanding of diverse real-world tasks. Another part is from EPIC-KITCHENS~\cite{damen2020epic} and RoboCerebra~\cite{han2025robocerebra}, which provides fine-grained sub-steps and time-segment annotations to supervise the learning of semantic progress estimation in long-horizon, contact-rich scenarios. To keep storage and computation requirements manageable during pre-training, the model predicts entire frames rather than fine-grained affordances. For the fine-tuning stage, we select 942 trajectories from robot data and annotate them with affordance data and continuous progress labels using a semi-automated method. We then fine-tuned the model on these annotated trajectories to learn conditional affordance foresight for inverse dynamics prediction, ultimately outputting an action-progress pair. For the post-training stage, we adopt dataset-specific schedules. For LIBERO~\cite{liu2023libero}, we pre-train on LIBERO-90, which contains fully annotated demonstrations for 90 short-horizon tasks, and then fine-tune and evaluate on LIBERO-LONG, which features long-horizon tasks. For CALVIN ABC$\rightarrow$D~\cite{mees2022calvin}, we first pre-train on the official robot play data without language instructions; the remaining language-conditioned data with full annotations is used for fine-tuning.

\noindent \textbf{Progress Labels.} We manually annotate functional keyframes per trajectory (\eg, \textit{Grasp-Contact}, \textit{Release}) and assign fixed semantic progress anchors; intermediate frames are linearly interpolated. In video pre-training, progress is assigned from \textit{observable interaction milestones} using the same anchors as in robot trajectories. Since these milestones are shared across human video and robot rollouts, they define a unified scale that ensures progress labels remain reliable during transfer, allowing PALM to disambiguate task stages regardless of domain or horizon length.

\noindent \textbf{Hyperparameters.} We perform training on 8 NVIDIA A100 GPUs, and set an initial learning rate of 1e-3, a weight decay of 1e-4, and a batch size of 64. Throughout the entire training process, the pre-trained visual and text encoders are kept frozen. All models are trained for a total of 30 epochs. Our model has 68 million trainable parameters. This lightweight architecture offers a compact and flexible base for fine-tuning, allowing it to be performed on smaller GPUs such as the RTX 4090. Details are illustrated in Table \ref{appendix_hyper}.

\begin{table}[t!]
\centering
\caption{
\textbf{Training hyperparameters.}
} \label{appendix_hyper}
\vspace{-0.3cm}
\small
\begin{tabular}{lll}
\toprule
\textbf{Hyperparameters} & \textbf{Pre-training} & \textbf{Fine-tuning} \\
\midrule
Number of GPUs & 8 & 8 \\
Batch Size & 80 / GPU & 64 / GPU \\
Learning Rate & 1e-4 & 1e-3 \\
Weight Decay & 1e-4 & 1e-4 \\
Optimizer & AdamW & AdamW \\
Learning Rate Schedule & Cosine decay & Cosine decay \\
Training Epochs & 30 & 40 \\
Historical Sequence Length & 7 & 7 \\
Action Prediction Length & 3 & 3 \\
\bottomrule
\end{tabular}
\vspace{-0.3cm}
\end{table}

\subsection{Policy Roll-out Details}
To ensure efficiency during inference, we set the number of DiT diffusion steps to 10, the number of observation steps to 7, and the number of future prediction steps to 3. This configuration results in a rapid sampling time of approximately 40 ms. Our approach uses fine-grained affordances for conditional visual foresight, which avoids the need for explicit image decoding. This enables a closed-loop frequency of 10-15 Hz, with each decision cycle taking less than 80 ms.

\section{Long-Horizon Real-World Task Details}
\label{sec:realworld}

\subsection{Task Setup and Success Criteria}
We design a long-horizon manipulation task to evaluate the model's ability to follow complex, sequential instructions. The experimental setup consists of a UFACTORY xArm 6 robot arm equipped with a Gripper G2. Visual perception is provided by two RealSense D455 cameras: one mounted on the robot's wrist for an eye-in-hand perspective, and another positioned statically for a third-person, eye-on-base view. The task is driven by a single, high-level language instruction: \texttt{"Put the pineapple on the white plate, the grape on the white bowl, and the orange on the blue bowl."} The scene contains three toy fruits (pineapple, grape, orange), two white containers (a plate and a bowl), and one blue bowl. For each trial, the initial positions of all objects are randomized within a predefined workspace to test policy robustness.
The full task is considered a success only when all three fruits are correctly placed in their designated containers as specified by the instructions. The success criteria for each of the six subtasks are detailed below.

\noindent $\diamondsuit$ \textbf{Subtask 1. Pick up pineapple:} The robot must successfully grasp the toy pineapple from its initial randomized location. Success for this subtask is defined as the robot securely gripping the pineapple and lifting it clear of the table surface without dropping it. This initial step challenges the policy's ability to generalize its grasping strategy to objects with irregular shapes and textures, evaluating its local geometric reasoning for identifying stable contact points.

\noindent $\diamondsuit$ \textbf{Subtask 2. Place on white plate:} The robot must transport the grasped pineapple and place it onto the white plate. Success is achieved if the pineapple is fully supported by the plate and remains stable after the gripper retracts. This step tests both precise spatial targeting and the model's capacity for language-based disambiguation, as it must correctly identify the ``plate'' from two similar white containers.

\noindent $\diamondsuit$ \textbf{Subtask 3. Pick up grape:} The robot must successfully grasp the toy grape from its initial position. The success criteria are identical to Subtask 1. This action requires fine-grained motor control, evaluating the policy's precision when interacting with small objects where the margin for positional error is minimal.\looseness-1

\noindent $\diamondsuit$ \textbf{Subtask 4. Place on white bowl:} The robot must transport the grasped grape and place it inside the white bowl. Success is achieved if the grape is fully contained within the bowl after the gripper retracts. This subtask again evaluates language disambiguation (``bowl'' vs. ``plate'') and tests the model's understanding of different placement affordances, specifically placing an object \textit{into} a concave container versus \textit{onto} a flat surface.\looseness-1

\noindent $\diamondsuit$ \textbf{Subtask 5. Pick up orange:} The robot must successfully grasp the toy orange from its initial position. The success criteria are identical to Subtask 1. This action serves as a baseline evaluation of the model's core grasping capability on a simple, regular object shape.\looseness-1

\noindent $\diamondsuit$ \textbf{Subtask 6. Place on blue bowl:} The robot must transport the grasped orange and place it inside the blue bowl. Success is achieved if the orange is fully contained within the bowl after the gripper retracts. This explicitly tests the model's language grounding for object attributes, as it must correctly associate the color ``blue'' with the appropriate target container.\looseness-1

\subsection{Robustness of Progress Estimates}
To evaluate the reliability and physical grounding of the learned self-progress indicator $p_t$, we subject a representative subtask to three categories of dynamic perturbations injected twice per episode at random timesteps. Specifically, we test: \circnum{1} random object relocations, \ie, changing the target's pose or position mid-execution to require geometric reactivity; \circnum{2} unseen lighting, \ie, switching illumination to out-of-distribution conditions (\eg, sudden dimming or color-temperature shifts) to require perceptual invariance; and \circnum{3} visual distractions, \ie, introducing additional distractor objects to induce workspace clutter and occlusions. This analysis verifies that the progress estimate tracks the underlying task state rather than spurious visual correlations, supporting robust closed-loop behavior in unstructured environments. The details and results are as follows.

\begin{figure*}[t!]
    \centering
\includegraphics[width=0.8\linewidth]{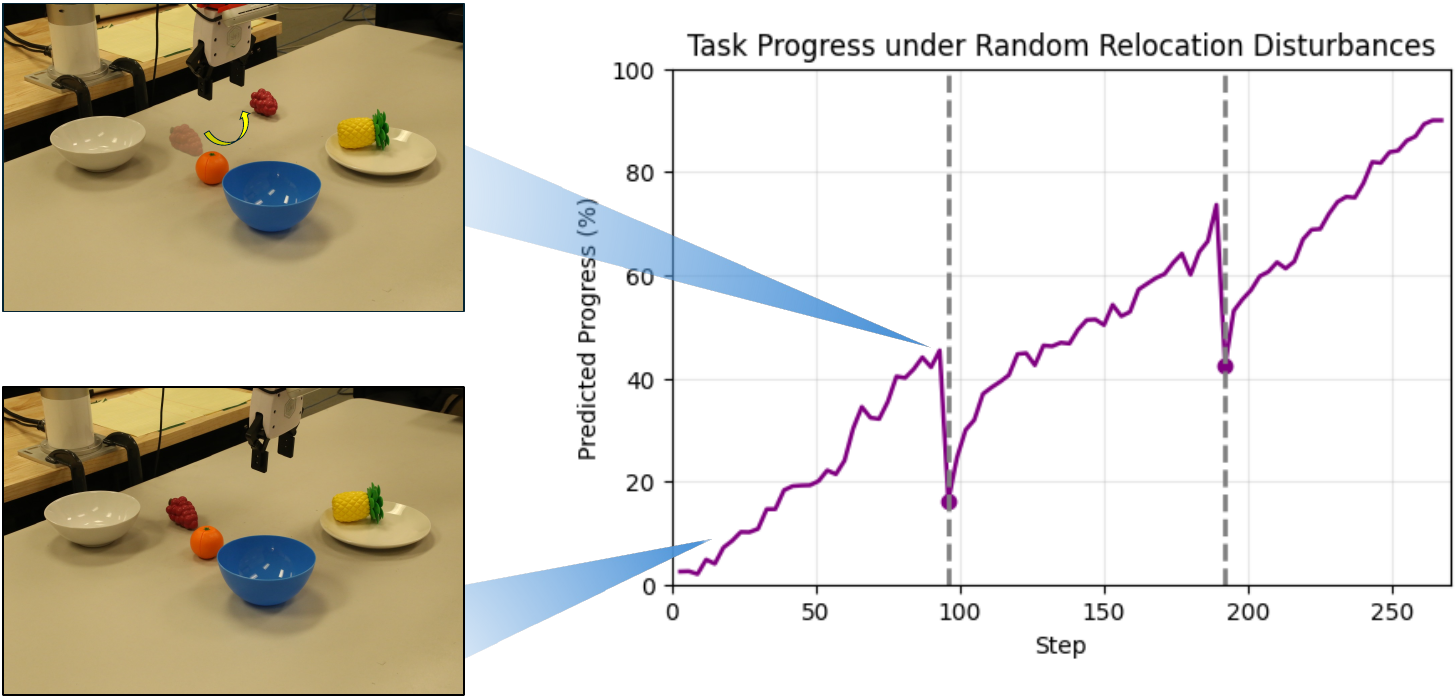}
    \vspace{-0.3cm}
    \caption{\textbf{Random Relocation Disturbances.} Predicted progress in the \texttt{"pick up grape"} subtask under two random grape relocations.}
    \vspace{-0.2cm}
    \label{fig:random_local}
\end{figure*}

\noindent $\diamondsuit$ \textbf{Random Relocation Disturbances.} As illustrated in \Cref{fig:random_local}, during the subtask of \texttt{"pick up grape"}, \emph{we relocate the grape to a new pose twice mid-execution} (vertical dashed lines). The predicted progress increases smoothly over time and, at each relocation event, exhibits a transient deviation followed by a rapid recovery that quickly returns to the previous upward trend rather than resetting or collapsing. This behavior indicates that the progress signal is robust to substantial changes in the target object’s pose and continues to track subtask completion rather than the instantaneous geometric configuration of the scene.

\begin{figure*}[t!]
    \centering
\includegraphics[width=0.8\linewidth]{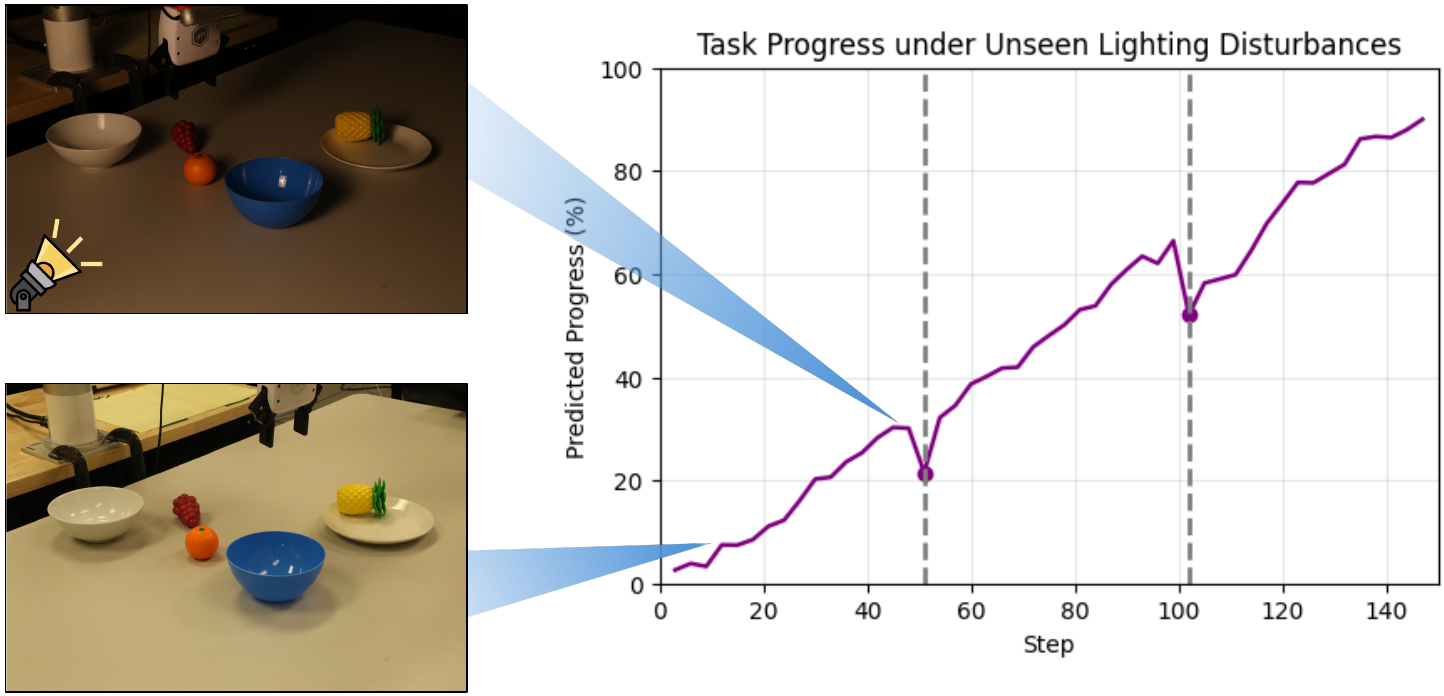}
    \vspace{-0.3cm}
    \caption{\textbf{Unseen Lighting Disturbances.} Predicted progress in the \texttt{"pick up grape"} subtask under two unseen lighting changes.}
    \vspace{-0.2cm}
    \label{fig:unseen_light}
\end{figure*}

\noindent $\diamondsuit$ \textbf{Unseen Lighting Disturbances.} As shown in \Cref{fig:unseen_light}, during the subtask of \texttt{"pick up grape"}, we introduce two abrupt switches to unseen illumination conditions (vertical dashed lines), substantially altering global brightness and shadows while leaving the physical scene unchanged. The predicted progress continues to increase over time, with only mild local deviations at the change points before returning to its prior upward trend. This indicates that the progress estimator is robust to severe photometric disturbances, exhibiting stable behavior under illumination changes.

\begin{figure*}[t!]
    \centering
\includegraphics[width=0.8\linewidth]{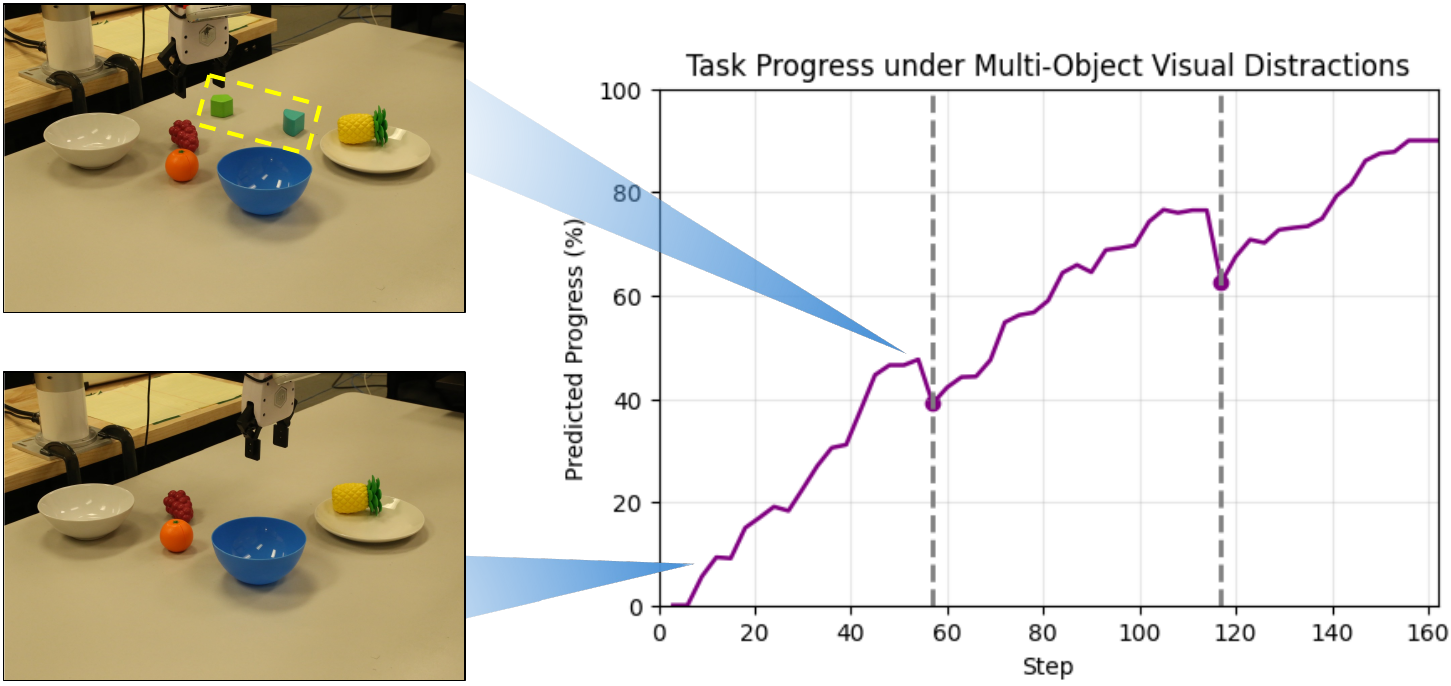}
    \vspace{-0.3cm}
    \caption{\textbf{Multi-Object Visual Distractions.} Predicted progress in \texttt{"pick up grape"} under two injected visual distraction events.}
    \vspace{-0.2cm}
    \label{fig:visual_distract}
\end{figure*}

\noindent $\diamondsuit$ \textbf{Multi-Object Visual Distractions.} As illustrated in \Cref{fig:visual_distract}, during the subtask of \texttt{"pick up grape"}, we introduce \emph{multiple additional objects into the scene twice} (vertical dashed lines), placing them near the target and within the camera’s field of view to create visual clutter. The predicted progress keeps increasing over time, with localized deviations at the distraction events that promptly recover and return to the overall upward trend. This indicates that the progress estimator remains stable under multi-object visual distractions, with limited sensitivity to distractors.

\section{More Experiments}
\label{sec:extra_exp}
\subsection{Ablation on Progress Threshold}
The threshold $\phi$ serves as the decision boundary for terminating the current sub-policy and triggering the subsequent phase based on the predicted progress signal $p_t$. Since $p_t$ serves as a continuous indicator of the sub-task's temporal progress, the setting of this threshold directly governs the transition timing: a lower threshold risks premature termination of the action, while an excessively high threshold may induce execution stagnation due to signal non-saturation. To investigate the impact of this parameter on final manipulation performance, we conduct an ablation study on the CALVIN ABC$\rightarrow$D benchmark across $\phi \in\{70 \%, 80 \%, 90 \%, 100 \%\}$, with $\phi=90 \%$ as the default setting. This analysis aims to validate the effectiveness of the switching logic in long-horizon tasks by quantifying this trade-off. 

As illustrated in Table~\ref{tab:ablation_progress}, the default threshold of $90 \%$ yields the highest success rates across all chain lengths, achieving an average sequence length of 4.48. Reducing $\phi$ to $70\%$ significantly degrades performance due to premature transitions, whereas increasing it to 100\% causes a slight decline, confirming that a strict saturation requirement can hinder task completion.\looseness-1

\begin{table}[t!]
\caption{\textbf{Ablation studies on the progress threshold $\phi$.} Results on the CALVIN ABC$\rightarrow$D benchmark demonstrate the effect of the threshold setting on long-horizon task performance.}
\vspace{-0.3cm}
\centering
\begingroup
\setlength{\tabcolsep}{3pt}  
\resizebox{\columnwidth}{!}{%
\begin{tabular}{ccccccc}
\toprule
\multirow{2}{*}{Threshold $\phi$} & 
\multicolumn{6}{c}{Task completed in a row} \\ \cmidrule{2-7} 
 & 1 & 2 & 3 & 4 & 5 & Avg. Len.  \\ 
\midrule 
70\%  & 93.4 & 86.3 & 78.4 & 71.9 & 65.5 & 3.96 \\
80\% & 95.3 & 90.2 & 84.2 & 79.7 & 74.1 & 4.24 \\
\rowcolor{gray!20}
\textbf{90\%}  & \textbf{96.9} & \textbf{93.8} & \textbf{89.3} & \textbf{85.9}  & \textbf{82.0}  & \textbf{4.48} \\
100\% & 95.2 & 89.7 & 84.4 & 78.3 & 73.0 & 4.21 \\
\bottomrule
\end{tabular}%
}
\label{tab:ablation_progress}
\endgroup
\end{table}

\begin{table}[t!]
\begingroup
\captionof{table}{\textbf{Ablation on prediction vs. reconstruction.} Results on CALVIN ABC$\rightarrow$D and inference latency, demonstrating the impact of different prediction and reconstruction objectives on long-horizon task performance.}
\vspace{-0.3cm}
\resizebox{\columnwidth}{!}{
\begin{tabular}{ccc}
\toprule
\multirow{2}{*}{Ablation Type} & 
CALVIN ABC$\rightarrow$D & 
Latency \\ \cmidrule(lr){2-2} \cmidrule(lr){3-3}
& Avg. Len. $\uparrow$ & (ms) $\downarrow$  \\ 
\midrule
\rowcolor{gray!20} \textbf{Affordance} \textbf{(\modelname)} & \textbf{4.48} & \textbf{$\sim$70} \\ 
Image / Video & 4.17  & $\sim$90 \\ 
Auxiliary & 3.58 & $\sim$55  \\ 
\bottomrule
\end{tabular}%
} 
\label{tab:prediction_type}
\endgroup
\vspace{-0.3cm}
\end{table}

\subsection{Ablation on Prediction vs. Reconstruction}
We perform an ablation on prediction targets vs. reconstruction to understand which form of foresight best supports long-horizon control. With the backbone and task-conditioned policy fixed, we compare: (1) Affordance Foresight, which predicts future affordance maps to highlight actionable regions and interaction points; (2) Image/Video Foresight, which predicts future RGB observations as a purely pixellevel forecasting objective; and (3) Auxiliary or Reconstruction, which reconstructs the current observation at time $t$ as a representation-learning signal without future prediction. All variants are evaluated on CALVIN ABC$\rightarrow$D using the average number of consecutively completed subtasks (Avg. Len.) and inference latency.

As shown in Table~\ref{tab:prediction_type}, affordance foresight attains the highest average length (4.48) with moderate latency ($\sim$70 ms), while future RGB prediction yields lower performance (4.17) at higher latency ($\sim$90 ms), and auxiliary reconstruction is fastest ($\sim$55 ms) but degrades performance the most (3.58). This pattern indicates that structured, action-centric affordance prediction yields substantial long-horizon gains without prohibitive computational cost, and provides the most effective trade-off among the tested prediction types.

\subsection{Analysis of Real-Robot Failure Modes}
Table~\ref{tab:real_robot_failure_modes_only} analyzes real-robot failures in the same 6-subtask setting with $N{=}50$ rollouts per method. We report episode-level \textbf{Repeat} (small ineffective motions), \textbf{Skip} (attempting later subtasks early), and \textbf{Premature Stop} (sustained near-zero motion before completion). Rates are the fraction of episodes with $\ge$1 event under identical detection rules. PALM reduces such failures compared to baselines.

\begin{table}[t]
\centering
\caption{\textbf{Quantifying real-robot failure modes.} Compared with baseline, PALM achieves longer executions and substantially reduces repeat, skip, and premature-stop errors on the real robot.}
\label{tab:real_robot_failure_modes_only}
\vspace{-0.3cm}
\resizebox{\columnwidth}{!}{
\begin{tabular}{lcccc}
\toprule
Method & Avg. Len. $\uparrow$ & Repeat $\downarrow$ & Skip $\downarrow$ & Premature \\ 
&  & Rate & Rate & Stop $\downarrow$ \\
\midrule
OpenVLA~\cite{kim2024openvla} & 2.30 & 28\% & 16\% & 22\% \\
Octo~\cite{team2024octo} & 1.85 & 34\% & 26\% & 18\% \\
\rowcolor{gray!20} \textbf{PALM} & \textbf{3.90} & \textbf{14\%} & \textbf{8\%} & \textbf{10\%} \\
\bottomrule
\end{tabular}
}
\vspace{-0.3cm}
\end{table}

\section{Qualitative Results and Visualization}
\label{sec:qualitative}

\begin{figure*}[t!]
    \centering
\includegraphics[width=0.98\linewidth]{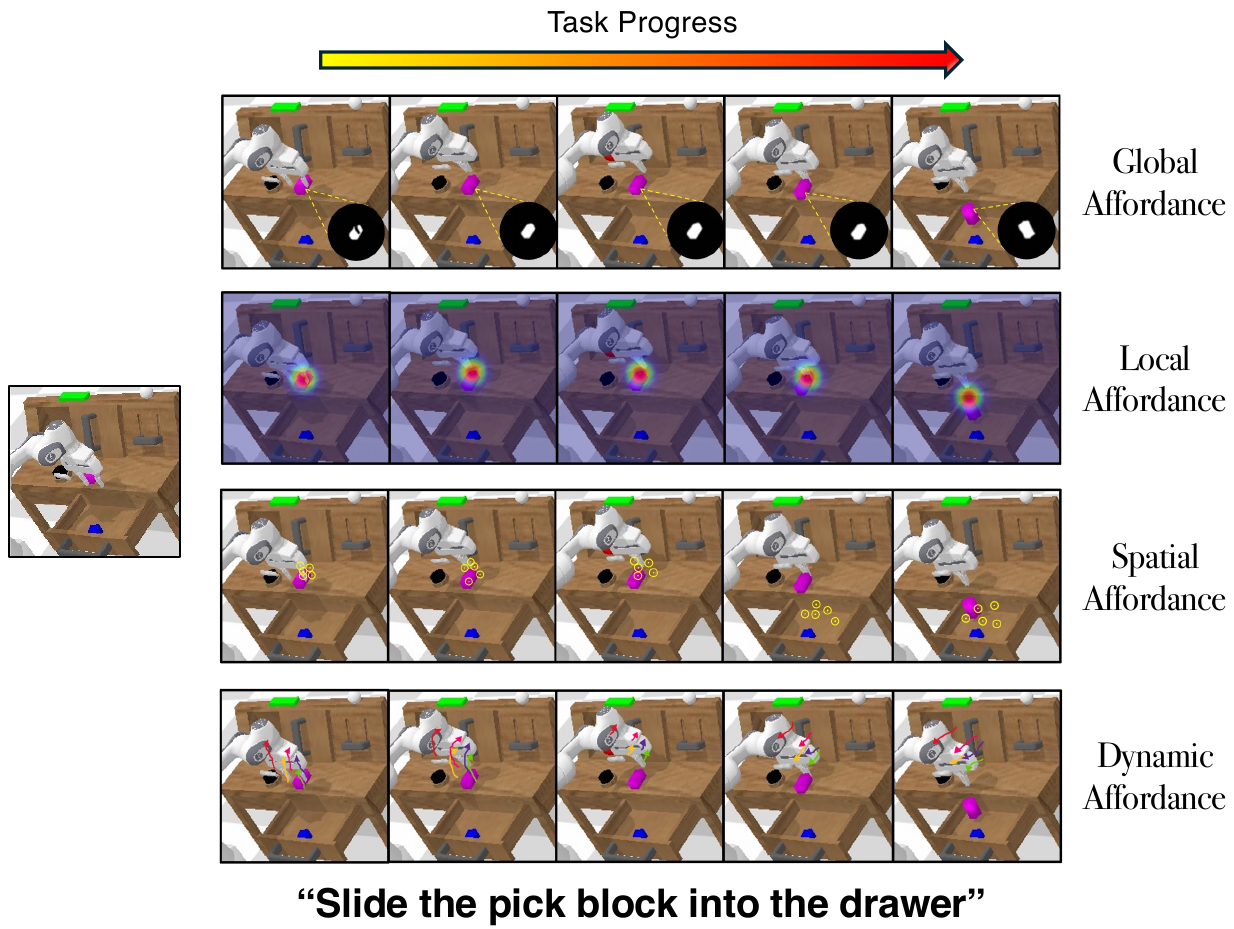}
    \vspace{-0.4cm}
    \caption{\textbf{Visualization of affordance predictions.} Across sequential progress steps, the model predicts four complementary affordances to guide policy generation: \textbf{Global Affordance} segments task-relevant objects and goals; \textbf{Local Affordance} generates heatmaps for precise contact points; \textbf{Spatial Affordance} predicts candidate placement regions; and \textbf{Dynamic Affordance} forecasts motion trajectories.}
\label{fig:calvin_visual}
    \vspace{-0.2cm}
\end{figure*}

In this section, we visualize the internal reasoning process of \modelnamenc{} during the execution of a long-horizon manipulation task: ``Slide the pick block into the drawer.'' As illustrated in Figure~\ref{fig:calvin_visual}, our model actively predicts structured affordances to guide its decision-making. We display the rollout over five timesteps, visualizing the four distinct affordance outputs that the model generates:

\begin{itemize}
    \item \textbf{Global Affordance:} The model correctly segments the target object (the pink block) and the goal region (the open drawer), demonstrating semantic understanding of the instruction.
    \item \textbf{Local Affordance:} The heatmaps focus precisely on the interaction points, shifting from the block's graspable surface to the handle of the drawer as the task progresses.
    \item \textbf{Spatial Affordance:} The predicted yellow keypoints indicate valid placement candidates, guiding the robot to move the block towards the drawer's opening.
    \item \textbf{Dynamic Affordance:} The arrows visualize the predicted motion of the end-effector and the object, showing a clear trajectory for sliding the block forward.
\end{itemize}

These visualizations confirm that \modelname builds a comprehensive, structured representation of the task. By explicitly forecasting \textit{what} to interact with (Global/Local), \textit{where} to move (Spatial), and \textit{how} the scene will evolve (Dynamic), the model achieves precise and robust control, successfully completing the task where baselines often fail due to ambiguity or lack of spatial reasoning.

\section{Broader Impacts}
\label{sec:impact}
\modelnamenc{} advances robotic manipulation by enhancing intermediate reasoning to mitigate common failures in long-horizon tasks. Rather than direct sensorimotor mapping, the model first anticipates a structured set of future affordances that encode task-relevant object identities, interaction points, spatial goal regions, and motion patterns. It then couples these predictions with a mechanism for estimating progress within the current subtask. This combination of structured affordance prediction and progress-aware state tracking yields an internal representation that helps policies remain coherent and effective across complex multi-stage activities where traditional models often fail.
At the application level, progress-aware, affordance-based policies can help move long-horizon manipulation from controlled labs to open-world settings (\eg, homes, warehouses, hospitals) when paired with sufficient data and stronger inductive priors. Because \modelnamenc{} exposes explicit affordance and progress signals, it integrates more naturally with safety monitors, constraint checks, and human-in-the-loop control, and it simplifies visualization and debugging relative to black-box policies.\looseness-1

However, these structured representations may reflect dataset biases (\eg, which objects are prioritized or what counts as success), and premature deployment in safety-critical environments could amplify such risks. Realizing the benefits, therefore, requires standardized evaluation, robust sim-to-real transfer, and careful safety and societal-impact assessments so that capability gains do not come at the expense of safety, employment, or privacy.

\section{Limitations and Future Work}
\label{sec:impact}
While \modelnamenc{} has significantly improved long-horizon performance, its online recovery capability remains limited under execution drift caused by partial observability, contact uncertainty, occlusions, and geometric deformations. We plan to improve deviation detection and state-consistency checks to obtain more accurate state estimates and, after localizing drift sources at the subtask, object, and contact levels, trigger replanning or robust rollback based on the current state. At the same time, affordance-oriented segmentation, state grounding, and VLM-based semantic interpretation still introduce nontrivial overhead in annotation, perception, and inference~\cite{wang2025care,du2026unsupervised,lai2025seer,li2025maris,zhang2025cross,pan2026frequency,yin2025semi,li2025exploring,li2025stitchfusion}. We further plan to combine this task-conditioned control framework with reinforcement learning (RL) and failure analysis from experience~\cite{lu2025robofac,xiao2025self,intelligence2025pi06vlalearnsexperience,gu2025safe,zhang2025reinbot,wang2024eaco}, while extending it to 3D world perception~\cite{huang2026pointworld,zhang2025vibes,zhang2024hoi,yu2025core3d,zhao2024m,chen2025language}, to further improve execution stability, robustness, and scalability for broader industrial applications~\cite{ye2026world,li2026causal,kim2026cosmos,yuan2026fast,yang2025drivemoe,yang2025raw2drive,yang2023llm4drive}.

\end{document}